\documentclass[10pt,twocolumn,letterpaper]{article}
\usepackage[pagenumbers]{cvpr}
\input{glyphtounicode}
\usepackage{amsthm,tabularx}
\usepackage{algorithm,algpseudocode,float,placeins,needspace}
\usepackage{colortbl,tikz}
\usetikzlibrary{arrows.meta,positioning}
\definecolor{traceRow}{RGB}{235,247,246}

\definecolor{cvprblue}{rgb}{0.21,0.49,0.74}
\usepackage[pagebackref,breaklinks,colorlinks,allcolors=cvprblue]{hyperref}
\newtheorem{proposition}{Proposition}

\newcommand{\Ans}{\mathcal A}

\title{WINE: Why Information Needs Evidence?}
\author{Sanghong Kim, Namkyung Kim, Keyunghyun Yoo, Hwangnam Kim\\
Korea University\\
145, Anam-ro, Seongbuk-gu, Seoul, 02841, Republic of Korea\\
{\tt\small \{sanghongkim, nkyoon93, seven1705, hnkim\}@korea.ac.kr}
}
\hypersetup{pdftitle={WINE: Why Information Needs Evidence?}, pdfauthor={Sanghong Kim, Namkyung Kim, Keyunghyun Yoo, Hwangnam Kim}}
\begin{document}
\maketitle
\begin{abstract}
Useful perception depends on whether unresolved evidence can affect the answer. We propose WINE, a structured vision-language model (VLM) cascade that uses answer sufficiency to coordinate observation, local answering, and escalation. Given supplied scene structure and a question program, supervised small-VLM heads produce attribute candidates; unread attributes retain full domains. A shared execution trace checks answer agreement and ranks reads by intermediate ambiguity, permitting local answers from partial perception without separate correctness gate training. Local correctness requires correct structure, a faithful defined program, and truth-containing evidence. On supported GQA questions, WINE achieves 78.29\% accuracy with 55.50\% fewer calls than image-and-question-only 8B inference. At matched 8B usage, it outperforms both tested attribute- and answer-confidence gates. Across three backbones, intermediate-state selection improves CLEVR accuracy over final-answer-set selection under half-scene attribute-read caps with shared perception, execution, and stopping. A single-GPU live GQA comparison also shows lower mean latency and calculated model FLOPs relative to 8B alone.
\end{abstract}

\providecommand{\PENDING}[1]{\textcolor{red}{[PENDING: #1]}}
\section{Introduction}
\label{sec:introduction}

Visual questions about the same image can require different amounts of evidence. A local attribute may determine one answer, while another requires resolving several objects and their relations. Efficient inference must therefore decide how much to observe before answering. Premature escalation and exhaustive observation can both incur avoidable computation. Model cascades route questions between models \citep{gupta2024language}, and small VLMs make local image processing practical \citep{marafioti2025smolvlm}. Combining them introduces a control problem: deciding when local evidence already suffices, which observation to acquire when it does not, and when to invoke a larger model.

Confidence gates and supervised routers guide escalation using perceptual or answer-level signals \citep{rabanser2025gatekeeper,ong2025routellm}. Confidence measures how uncertain perception is, not whether that uncertainty can change the answer. In compositional questions \citep{hudson2019gqa}, resolving an object reference can enable a later query before the final answer set changes. Conversely, some attributes can remain uncertain once all retained alternatives imply the same answer. Ambiguous references still permit material answers when all candidate objects are metallic. These two cases connect acquisition and stopping: the controller must track both progress within the reasoning process and agreement at its output.

\begin{figure*}[!t]
\centering
\includegraphics[width=\linewidth]{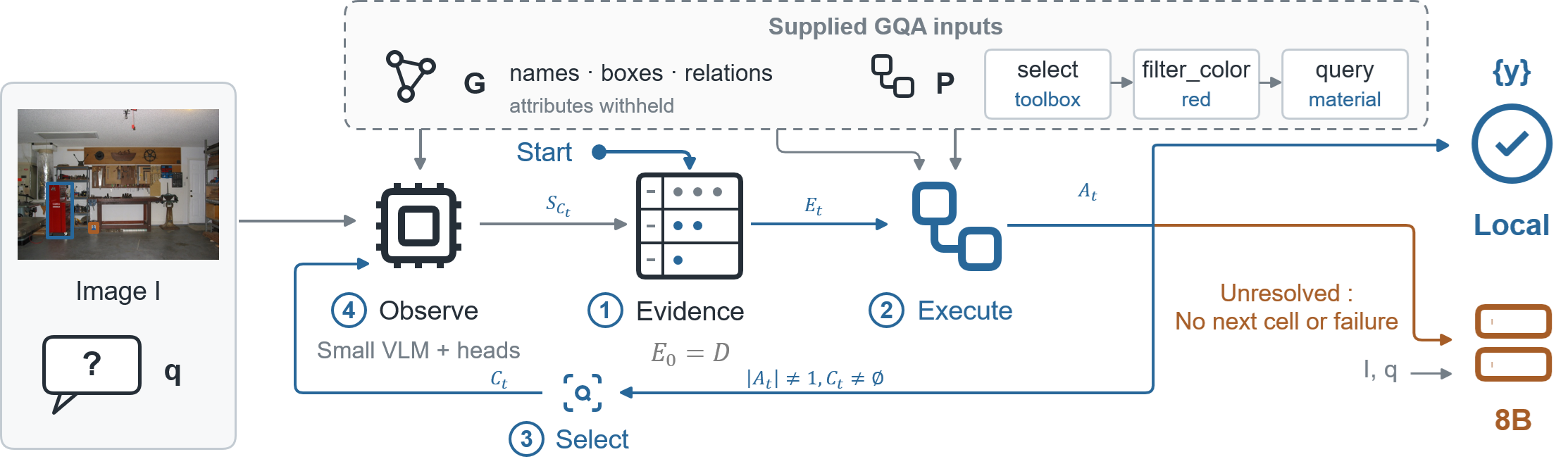}
\caption{WINE architecture with supplied GQA structure and an example question program. Numbered stages denote initialization, execution, selection, and observation.}
\label{fig:overview}
\end{figure*}

We propose WINE, an evidence-sufficiency controller for structured VLM cascades. Given supplied scene structure and a question program, supervised heads on a frozen small VLM provide attribute candidates; unread attributes retain full domains. By executing before perception is complete, the controller tracks intermediate references and possible answers in a set-valued execution trace. It checks answer agreement for local stopping and ranks hypothetical reads by intermediate ambiguity. As illustrated in \cref{fig:overview}, each observation updates this state; failed or unresolved local processing invokes Qwen3-VL-8B. WINE thus couples evidence acquisition and answer sufficiency without a separately trained correctness gate.

WINE reaches 78.29\% GQA accuracy with 55.50\% fewer 8B calls. Table~\ref{tab:routing-rules} shows that sufficiency routing outperforms the tested attribute- and answer-confidence gates at matched 8B usage. Attribute observations expand local answering beyond the Zero-read + 8B control, isolating coverage gained after observation from answers already fixed by supplied structure. CLEVR comparisons isolate the value of intermediate-state scoring under half-scene attribute-read caps with perception, execution, and stopping fixed. A separate live GQA comparison reduces mean latency by 14.81\%.

Our contributions are:
\begin{itemize}
\item We establish an answer-sufficiency criterion for incomplete attribute evidence, with conditional correctness under explicit structural, program, and evidence assumptions.
\item We rank hypothetical reads by intermediate-state ambiguity without encoding unread cells, valuing reference resolution before the final answer set changes.
\item We integrate the criterion and acquisition rule in a shared-trace controller that coordinates observation, local answering, and escalation without separate correctness gate training.
\end{itemize}

\section{Related work}
\label{sec:context}

\noindent\textbf{Compositional execution and uncertainty.} Neuro-symbolic VQA executes question programs over structured scene representations \citep{yi2018neural,mao2019neuro}, while VISPROG and ViperGPT generate programs that compose pretrained modules \citep{gupta2023visual,suris2023vipergpt}. These decompositions expose perceptual dependencies that WINE uses to guide observation, propagating alternative attribute values through supplied programs. This representation draws on abstract interpretation and possible-world semantics \citep{cousot1977abstract,imielinski1984incomplete}. Compositional conformal prediction also propagates sets through neurosymbolic programs, using calibration to obtain output coverage guarantees \citep{ramalingam2024conformal}. WINE executes over incomplete attributes using fixed-mass candidates; answer agreement controls conditional stopping, while intermediate states guide acquisition. Section~\ref{sec:selection-ablation} compares the official NS-VQA executor under shared perception and supplied programs, with implementation details in Appendix~\ref{app:published}.

\smallskip
\noindent\textbf{Active evidence acquisition.} Active feature acquisition trades observation cost against prediction quality. Joint acquisition and classification learns a predictor and a policy with observation and stopping actions \citep{shim2018joint}; EDDI selects features by expected information gain under a learned generative model \citep{ma2019eddi}. ReCoVERR gathers visual evidence through follow-up questions and checks entailment of an initially proposed answer \citep{srinivasan2024recoverr}. WINE exploits the supplied program to restrict observations to object--attribute cells and ranks hypothetical reads by residual execution ambiguity. Intermediate reference resolution can thereby guide acquisition before the final answer set changes. The score uses hypothetical symbolic updates, so candidate observations can be compared before their crops are encoded.

\smallskip
\noindent\textbf{Selective computation and routing.} Selective VQA estimates answer correctness to balance coverage and risk \citep{whitehead2022reliable,dancette2023improving}, while learning to defer trains a classifier and rejector using expert outcomes \citep{mozannar2020consistent}. Language-model cascades and routers allocate inference using answer scores, token uncertainty, preferences, or confidence tuning \citep{chen2024frugalgpt,gupta2024language,ong2025routellm,rabanser2025gatekeeper}. PARALLEL \citep{yoon2026parallel} uses goal- and uncertainty-related signals with utility--cost feedback to allocate sample-dependent parameter updates during adaptation. WINE reuses execution state for local acceptance and observation selection, unifying answer sufficiency and acquisition. Section~\ref{sec:routing} compares confidence gates, supervised routers, and an adapted published Selector with shared answers.

\section{Design of WINE}
\label{sec:method}
Execution over incomplete evidence supplies two control signals: final-answer agreement for stopping and intermediate ambiguity for acquisition. Figure~\ref{fig:overview} shows how the shared trace links perception and model escalation.

\subsection{Problem formulation}
\label{sec:inputs}
We consider structured visual question answering with selective perception. An instance comprises an image $I$, a question $q$, supplied scene structure $G$, and a typed program $P=(p_1,\ldots,p_m)$ intended to encode $q$. The structure specifies object identities, localization, and relations; program operations specify attribute predicates and dependencies between intermediate results. Object attributes remain unobserved until acquired from the image. In the primary GQA setting, $G$ contains object names, bounding boxes, and directed relation edges, while $P$ is obtained from the semantic annotations provided with the dataset. Input preparation is specified in Section~\ref{sec:setup} and Appendix~\ref{app:protocols}.

An attribute cell $c=(o,a)$ associates object $o$ with attribute $a$ and finite domain $D_a$. After $t$ actual reads, $E_t(c)\subseteq D_a$ denotes the nonempty set of retained values for cell $c$ and $Q_t$ the observed cells. Stage~1 in \cref{fig:overview} sets $E_0(c)=D_a$ and $Q_0=\varnothing$; unread cells retain full domains. The observed-cell set $Q_t$ distinguishes unread cells from observed cells whose candidates still span the full domain. The attribute reader comprises a frozen small VLM and supervised heads, predicting color and material in the primary cascade. Execution, acquisition, and stopping follow fixed rules. WINE returns either a local program answer $y$ or the answer from Qwen3-VL-8B conditioned only on $(I,q)$.

\subsection{Set-valued program execution}
\label{sec:execution}
Set-valued execution tracks the consequences of attribute alternatives through each program operation. At stage~2, a successful execution of $P$ over $(G,E_t)$ produces intermediate states $z_{t,1},\ldots,z_{t,m}$ and a final answer set $\Ans_t$. Object sets use mandatory and possible members $(M,U)$ with $M\subseteq U$; single-object references use candidate IDs $C$; scalar and Boolean results use finite sets $V$. These typed states preserve alternatives from the cell-level evidence through to the answer.

For a positive attribute filter with value $v$, the transfer rule is
\begin{equation}
 \begin{aligned}
 M'&=\{o\in M:E_t(o,a)=\{v\}\},\\
 U'&=\{o\in U:v\in E_t(o,a)\}.
 \end{aligned}
 \label{eq:filter}
\end{equation}
where $E_t(o,a)$ abbreviates $E_t((o,a))$. The resulting state encloses concrete sets between $M'$ and $U'$. An existence query is therefore true when $M'$ is nonempty, irrespective of uncertainty in the remaining objects, and false when $U'=\varnothing$. Single-object operations take $C=U$ and fail when $U=\varnothing$ or $|M|>1$, because these states violate the uniqueness requirement for a retained scene. A failed operation invalidates the current execution.

An attribute query over a reference $C$ returns the union of the candidate values for its objects. Consequently, an uncertain referent can still yield a definite attribute when all candidate objects share that value. Boolean conjunction and disjunction likewise combine the retained truth values of their operands. These transfers preserve answer alternatives without committing to a single perceived scene; Appendix~\ref{app:execution} gives the remaining query and comparison rules.

The answer set supports the sufficiency test below; intermediate states guide acquisition in Section~\ref{sec:selection}. Section~\ref{sec:stopping} connects these signals to escalation.

\begin{figure*}[!t]
\centering
\includegraphics[width=0.90\linewidth]{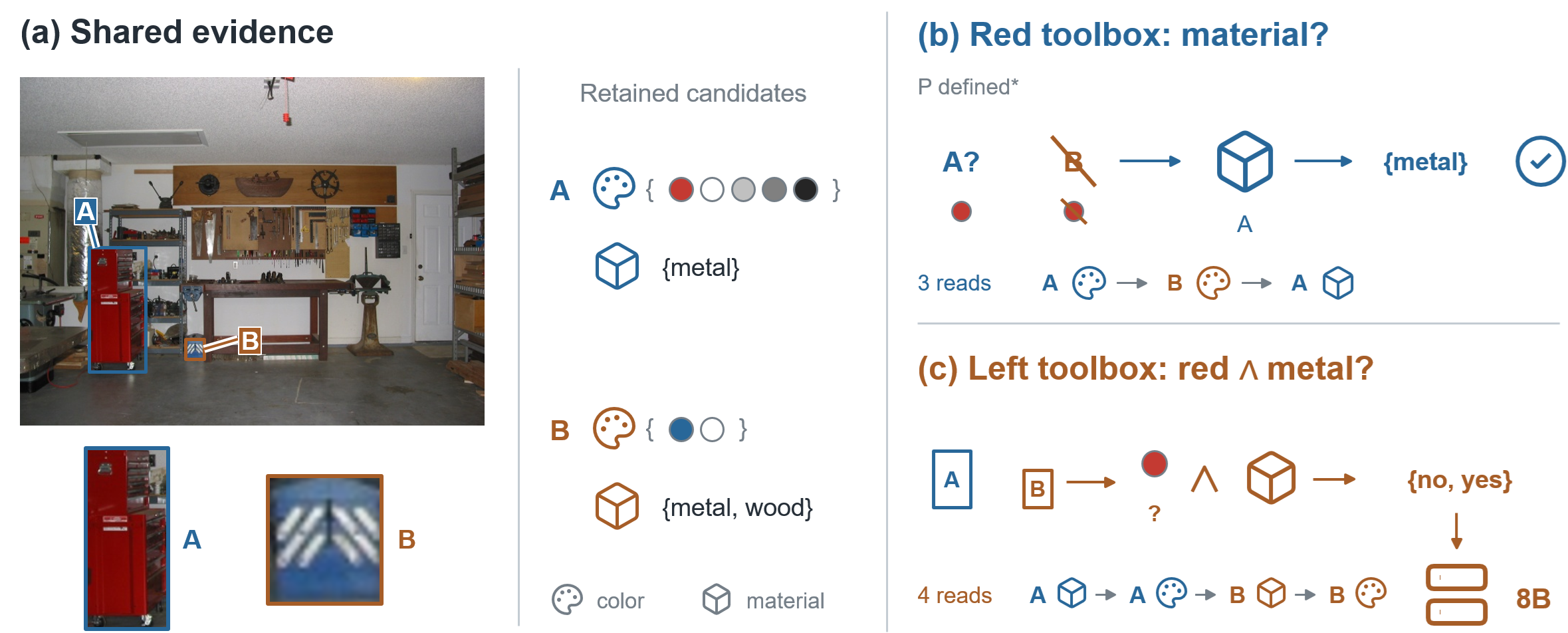}
\caption{Program-dependent evidence sufficiency. Panel (b) assumes exactly one red toolbox (*); panel (c) tests redness explicitly. Shared candidate evidence (a) therefore supports different stopping decisions.}
\label{fig:example}
\end{figure*}

\subsection{Conditional answer sufficiency}
\label{sec:guarantee}
Answer containment gives a reason to stop before every attribute is resolved. A program is defined on a scene if every concrete operation has a valid result, including single-object uniqueness requirements. Let $\Gamma(E_t,G)$ denote the scenes consistent with the evidence and supplied structure, and $\operatorname{Exec}(P,S,G)$ the concrete program answer on scene $S$. On successful execution, the supported typed executor satisfies:
\begin{equation}
 \begin{aligned}
 \{&\operatorname{Exec}(P,S,G):S\in\Gamma(E_t,G),\\
   &P\text{ defined on }S\}\subseteq\Ans_t.
 \end{aligned}
 \label{eq:containment}
\end{equation}
The abstraction may additionally retain infeasible answers because it discards dependencies between intermediate alternatives. Nevertheless, a singleton suffices for local correctness under the following premises: (A1) $G$ agrees with the true scene; (A2) $P$ is defined on that scene; and (A3) every $E_t(c)$, including unread domains, contains the true attribute value. Premises (A1) and (A3) place the true scene in $\Gamma(E_t,G)$.
\begin{proposition}[Conditional answer sufficiency]
\label{prop:sufficiency}
If the true scene belongs to \(\Gamma(E_t,G)\), \(P\) is defined on that scene, and
\(\Ans_t=\{y\}\), then \(y\) is the true program answer.
\end{proposition}
Appendix~\ref{app:execution} proves containment by induction over typed operations. Correctness for $q$ additionally assumes that $P$ faithfully encodes $q$; type validity checks operator compatibility. Given faithful inputs and truth-containing evidence, a singleton permits local answering while individual attributes remain unresolved, independently of read order.

\subsection{Trace-guided evidence acquisition}
\label{sec:selection}
Narrowing an intermediate reference can enable a later query even while the final answer set remains unchanged. At stage~3, WINE therefore ranks unread cells using intermediate-state ambiguity after a successful nonsingleton execution. The candidate pool $\mathcal C_t$ pairs objects appearing in any intermediate object set or reference with attributes used in $P$, restricted to supported cells absent from $Q_t$. This trace-wide pool retains cells even when later operations exclude their objects from the final referent.

Acquisition uses the aggregate ambiguity $\Phi(E_t;P,G)=\sum_{j=1}^m u(z_{t,j})$, with unit weight for each program node, including the final node. For object sets, references, and values, $u$ is respectively $|U|-|M|$, $|C|-1$, and $|V|-1$. This score counts alternatives retained along the trace.

For each $c\in\mathcal C_t$, hypothetical execution replaces $E_t(c)$ with a singleton $\{v\}$. Let $F_t(c)$ denote the values $v\in E_t(c)$ for which this execution succeeds. WINE selects the cell with minimum mean residual ambiguity:
\begin{equation}
 \begin{aligned}
 R_t(c)&=\frac{1}{|F_t(c)|}\sum_{v\in F_t(c)}
       \Phi(E_t[c\leftarrow\{v\}];P,G),\\
 c_t&\in\arg\min_{c\in\mathcal C_t:\,F_t(c)\ne\varnothing}R_t(c).
 \end{aligned}
 \label{eq:selection}
\end{equation}
The uniform average assigns equal weight to each successful candidate value. Failed probes are excluded from the score without modifying actual evidence. If $F_t(c)=\varnothing$ for every candidate cell, the first pool cell is selected. Algorithm~\ref{alg:trace} in Appendix~\ref{app:execution} specifies the fixed tie order used in GQA. The score heuristically orders reads; containment in Section~\ref{sec:guarantee} governs acceptance independently. Both use retained alternatives, allowing reference resolution to receive priority before final-answer ambiguity decreases.

\paragraph{Attribute observation.}
\label{sec:observation}
At stage~4, observation of the selected cell $c_t=(o,a)$ produces a candidate set $S_{c_t}\subseteq D_a$. The GQA reader encodes the supplied object crop with an object-name/attribute prompt; a supervised head maps the frozen small-VLM feature to probabilities over $D_a$. The set $S_{c_t}$ is the smallest highest-probability prefix with cumulative mass at least $\tau=0.8$, fixed across the reported GQA evaluations. Appendix~\ref{app:protocols} specifies crop construction, head training, the empirical threshold rationale, and the separate diagnostic calibration.

The observation updates $E_{t+1}=E_t[c_t\leftarrow S_{c_t}]$ and $Q_{t+1}=Q_t\cup\{c_t\}$, leaving all other cells unchanged. Unlike a hypothetical singleton probe, $S_{c_t}$ may retain multiple values. Each cell is acquired at most once, and the updated evidence is re-executed before any further acquisition. Primary GQA accuracy evaluation reveals cached candidate sets at this step; the live benchmark computes them on demand under the same controller, as specified in Section~\ref{sec:setup}.

\subsection{Selective prediction and escalation}
\label{sec:stopping}
WINE checks sufficiency before the first perceptual call and after each acquired cell. An actual execution failure triggers 8B escalation. Otherwise, $|\Ans_t|=1$ returns the sole candidate $y$ locally; a successful nonsingleton permits acquisition only while $\mathcal C_t\neq\varnothing$ and the read cap has not been reached. This ordering allows zero-read returns from $G,P$ and retains the singleton check after the final permitted observation. Algorithm~\ref{alg:trace} gives the complete local controller.

The primary cascade imposes no read cap. An empty acquisition pool or budget exhaustion without a singleton invokes 8B on the original $(I,q)$; the larger model receives only the image and question. Standalone acquisition experiments omit 8B fallback and count unresolved outputs as abstentions.

\paragraph{Qualitative analysis.}
\Cref{fig:example} contrasts two questions sharing the candidate evidence in panel~(a). Panel~(b) asks for the material of the red toolbox. The asterisk marks a program premise: exactly one toolbox is red. The color candidates for A include red and other colors, whereas those for B exclude red. Among scenes satisfying the premise, A is therefore the referent, and its singleton material set yields \emph{metal} after three reads. The unique red referent follows from the program premise together with exclusion of B; the color observation for A remains ambiguous.

Panel~(c) identifies A by a spatial relation and tests whether it is red and metallic. Reading the material and color of A leaves $\{\mathrm{no},\mathrm{yes}\}$ because its color remains ambiguous. B remains in the initial \texttt{select} state, so its unread material and color cells remain eligible under the trace-wide pool in Section~\ref{sec:selection}. Acquisition continues while the answer is unresolved and the pool is nonempty, without requiring a positive ambiguity reduction. The controller therefore reads both cells of B, which leave the answer unchanged, before pool exhaustion triggers 8B escalation after four reads. This contrast illustrates program-dependent stopping: the material answer is sufficient in panel~(b), whereas pool exhaustion governs escalation in panel~(c).

\section{Evaluation}
\label{sec:evaluation}
Evaluation tests the two control roles of the shared trace. GQA measures how acquired evidence expands local answering beyond structure-fixed answers. CLEVR isolates intermediate-state acquisition with perception, execution, and stopping held fixed. Routing and component comparisons characterize acceptance without gate training; live inference measures whole-cascade cost. Relation and program edits test the premises of local stopping.

\subsection{Experimental protocol}
\label{sec:setup}

\paragraph{Data and structure.}
Primary GQA evaluation \citep{hudson2019gqa} retains supported color/material programs reproducing annotated answers on annotated graphs: 15.34\% of balanced validation questions. The holdout contains 10,509 questions on 3,738 training-split images; router fitting uses 15,714 questions on disjoint images. $G$ supplies names, boxes, and relations with attributes withheld; a fixed parser constructs $P$ from semantic annotations. Appendix~\ref{app:protocols} specifies preparation and partitions. Head and router fitting are image-disjoint from evaluation; SmolVLM pretraining includes GQA through The Cauldron \citep{laurencon2024idefics2}.

Color and material heads use frozen SmolVLM-256M features and 25,569 color and 3,026 material labels from images disjoint from routing panels. Five fits train both heads with different initializations. The 8B alone baseline invokes Qwen3-VL-8B \citep{bai2025qwen3vl} once on $(I,q)$. Primary evaluation reveals saved attribute predictions and reuses saved 8B answers; unread cells retain full domains. Live inference computes features, head outputs, and 8B answers on demand using the same controller. Head training is offline.

The live comparison uses 256 GQA questions and one fit. Routers make 118 8B calls per repetition, matched by panel-level ranking with the original MLP features and hyperparameters. Both backbones remain on one RTX 3090. Timing includes image processing, model calls, CPU heads, execution, and routing. Supervised MLP uses all 24 answer and execution features; Structure-first computes local answers only for retained questions. Appendix~\ref{app:matched-runtime} specifies weighting and timing.

\paragraph{Comparisons and metrics.}
Routing comparisons fix local and 8B answers and match WINE calls within each fit and panel, testing which questions to escalate. The shared local candidate is the WINE singleton or, otherwise, the modal answer from 256 sampled executions; WINE escalates unresolved cases. Acquisition comparisons fix perception and execution under a common read cap. Accuracy counts unresolved answers as errors. Local coverage divides returns by all questions; local risk divides wrong returns by local returns. Wrong local / all divides wrong returns by all questions. Throughout, $\Delta$ errors is reference errors minus WINE errors; positive values favor WINE.

Primary GQA routing results average five fits. Paired 95\% intervals use 10,000 bootstrap draws crossing fits and image clusters, resampling router seeds within fit. Read-budget and predicted-input panels use one fit; a separate 256-question CLEVR panel uses three fits per SmolVLM backbone. Live timing uses fit 0 and two repetitions. Appendix~\ref{app:comparisons} separately specifies the published-component and controlled-corruption studies, including their fixed-fit, 2,000-draw image-cluster intervals.

\begin{table}[!t]
\centering\small
\caption{Complete-system accuracy and 8B usage on 10,509 GQA questions (five-fit means).}
\label{tab:gqa-main}
\setlength{\tabcolsep}{3pt}
\begin{tabular*}{\columnwidth}{@{\extracolsep{\fill}}lrrr@{}}
\toprule
System & \shortstack{8B calls\\(\%)} & Acc. (\%) $\uparrow$ & Errors\,$\downarrow$\\
\midrule
Local answer only & 0.00 & 71.57 & 2,987.8\\
8B alone & 100.00 & 76.45 & 2,475.0\\
Zero-read + 8B & 93.62 & 78.10 & 2,301.0\\
\textbf{WINE+8B cascade} & 44.50 & 78.29 & 2,281.0\\
\bottomrule
\end{tabular*}
\end{table}

\subsection{Results}
\label{sec:results}
\subsubsection{Local answering from partial evidence}
\label{sec:gqa}
WINE extends local answering to questions resolved by visual evidence. Table~\ref{tab:gqa-main} compares it with Zero-read + 8B, which accepts answers fixed by $G,P$ and delegates the rest. Both cascades share the supplied structure and fallback answers, isolating the additional local returns enabled by attribute acquisition. 8B alone receives $(I,q)$; local-only returns the shared execution-derived candidate.

\begin{figure*}[!t]
\centering
\includegraphics[width=0.88\linewidth]{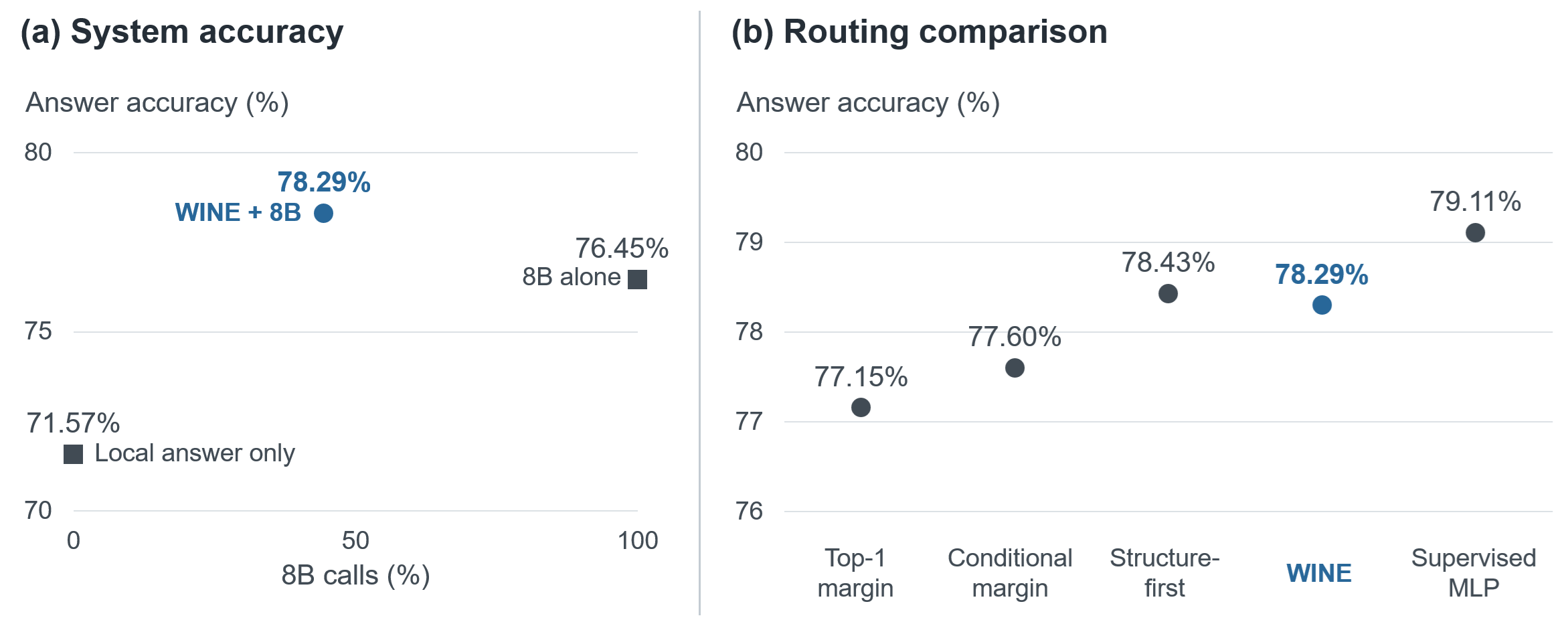}
\caption{System accuracy and 8B usage (a); routing accuracy at a matched 44.50\% mean call rate (b). Five-fit means.}
\label{fig:accuracy-calls}
\end{figure*}

WINE attains 78.29\% accuracy versus 76.45\% for 8B alone, with $\Delta$ errors of $+194$ $[132,254]$. Zero-read + 8B attains 78.10\% at 93.62\% 8B usage. WINE answers 55.50\% locally, reducing 8B usage to 44.50\%, with a 0.19-point accuracy difference from the zero-read control. Figure~\ref{fig:accuracy-calls}(a) plots system accuracy and 8B usage.

\begin{table}[!t]
\centering\small
\caption{$\Delta$ errors relative to 8B, grouped by zero-read acceptance.}
\label{tab:decision}
\setlength{\tabcolsep}{2.5pt}
\begin{tabular*}{\columnwidth}{@{\extracolsep{\fill}}lrrrr@{}}
\toprule
 & & \multicolumn{2}{c}{Final-answer errors} & \\
\cmidrule(lr){3-4}
Group & Count & 8B & WINE+8B & $\Delta$ errors [95\% CI]\\
\midrule
\shortstack[l]{Fixed by $G,P$\\(no read)} & 670 & 174 & 0 & $+174\;[147,202]$\\
\shortstack[l]{Not accepted\\before reads} & 9,839 & 2,301 & 2,281 & $+20\;[-34.8,75.8]$\\
\midrule
All & 10,509 & 2,475 & 2,281 & $+194\;[132,254]$\\
\bottomrule
\end{tabular*}
\end{table}

\label{sec:decomposition}
Table~\ref{tab:decision} separates 670 zero-read returns from 9,839 other questions, including subsequent local returns and 8B fallbacks. Relative to 8B alone, $\Delta$ errors is $+174$ $[147,202]$ and $+20$ $[-34.8,75.8]$, respectively. The split is identical across fits. Zero-read answers are correct under the supplied inputs and eligibility filter, accounting for most of the accuracy gain.

Across zero-read and after-read returns, local risk is 13.30\% for WINE and 16.63\% for 8B on the same questions, as detailed in Appendix~\ref{app:routers}. Checking answer agreement after each observation permits local returns while other attributes remain unresolved. Visual evidence can therefore replace a large-model call as soon as the retained alternatives determine an answer, expanding local coverage at close observed accuracy to the zero-read control.

\begin{table}[!t]
\centering\small
\caption{CLEVR accuracy (\%) under attribute-read budgets: 1,024 questions, SmolVLM-256M.}
\label{tab:read-budget}
\setlength{\tabcolsep}{4pt}
\begin{tabular*}{\columnwidth}{@{\extracolsep{\fill}}lrrr@{}}
\toprule
Read policy & 25\% & 50\% & 100\%\\
\midrule
WINE & 10.55 & 57.62 & 91.41\\
Final-set & 6.64 & 27.25 & 91.70\\
Fixed order & 1.27 & 7.91 & 91.21\\
Random order & 1.76 & 14.94 & 91.41\\
\bottomrule
\end{tabular*}
\end{table}

\subsubsection{Trace-guided attribute acquisition}
\label{sec:selection-ablation}
Table~\ref{tab:read-budget} tests read ordering on 1,024 CLEVR questions~\citep{johnson2017clevr} with supplied crop centers, relations, and programs. Methods share SmolVLM-256M heads, cached evidence, execution, unread pools, and stopping. WINE scores intermediate ambiguity; Final-set scores final-answer ambiguity. Fixed/random controls also change only ordering. Evidence is revealed after selection.

Each object has size, color, material, and shape cells. For $N$ objects, a fraction $f\in\{.25,.5,1\}$ permits $\lfloor4fN\rfloor$ observations from the reachable unread pool. The cap counts the full scene grid. Unresolved cases count as errors, with no 8B fallback. At the half-scene cap, WINE reaches 57.62\% versus 27.25\% for Final-set, a 30.37-point difference $[27.05,33.69]$; full-cap accuracies are 91.41\% and 91.70\%.

Across SmolVLM-256M, SmolVLM-500M, and Qwen3-VL-2B, half-cap WINE accuracy is 56.15--58.98\%, versus 26.37--27.93\% for Final-set. A separate 256-question CLEVR panel retains a 24.6--27.0-point advantage across three fits per SmolVLM backbone. On 896 CLOSURE questions from seven families~\citep{bahdanau2019closure}, the half-cap values are 25.89\% and 10.04\%. CLOSURE uses earlier CLEVR-trained attribute heads without refitting and supplied programs. Appendix~\ref{app:read-order} details the panels and bundled-observation diagnostic.

Intermediate-state guidance resolves more answers within the half-scene budget. Resolving a reference can advance execution before answer ambiguity decreases, giving acquisition a signal beyond the final answer set. Close full-cap accuracies locate the benefit in read ordering. The shared trace thus guides progress and detects sufficiency. Appendix~\ref{app:read-order} examines bundled crop observations; Section~\ref{sec:cost} measures cascade cost.

The supporting comparison in Table~\ref{tab:published-main} evaluates the official NS-VQA executor~\citep{yi2018neural} on these 1,024 questions with shared attribute heads. NS-VQA uses supplied relations and all argmax attributes; detection and parsing are excluded. WINE stops with candidate sets. Lower local risk accompanies lower local coverage, describing acceptance under the sufficiency rule. Appendix~\ref{app:published} reports observation counts and paired intervals for this acceptance comparison.

\begin{table}[!t]
\centering\small
\caption{GQA routing results at a matched mean 8B-call rate of 44.50\%.}
\label{tab:routing-rules}
\setlength{\tabcolsep}{2.5pt}
\begin{tabular*}{\columnwidth}{@{\extracolsep{\fill}}lrrr@{}}
\toprule
Routing method & \shortstack{Acc.\\(\%)} & Errors & \shortstack{$\Delta$ errors\\{[95\% CI]}}\\
\midrule
Top-1 margin & 77.15 & 2,401.0 & $+120.0\;[66.0,172.0]$\\
\begin{tabular}[c]{@{}l@{}}Conditional margin\\($K=1{,}024$)\end{tabular} & 77.60 & 2,354.4 & $+73.4\;[44.2,102.8]$\\
Structure-first & 78.43 & 2,267.2 & $-13.8\;[-63.6,34.4]$\\
WINE & 78.29 & 2,281.0 & ---\\
Supervised MLP & 79.11 & 2,195.8 & $-85.2\;[-117.6,-53.8]$\\
\bottomrule
\end{tabular*}
\end{table}

\begin{table}[!t]
\centering\small
\caption{Supporting component comparisons under shared inputs.}
\label{tab:published-main}
\setlength{\tabcolsep}{3pt}
\begin{tabular*}{\columnwidth}{@{\extracolsep{\fill}}lrrr@{}}
\toprule
Method & \shortstack{Acc.\\(\%)} & \shortstack{Local coverage\\(\%)} & \shortstack{Local risk\\(\%)}\\
\midrule
\multicolumn{4}{@{}l}{CLEVR: 1,024 questions, one fit, no 8B}\\
WINE & 91.41 & 91.89 & 0.53\\
NS-VQA (adapted) & 91.70 & 93.95 & 2.39\\
\midrule
\multicolumn{4}{@{}l}{GQA: fit 0, 45.87\% 8B calls}\\
\begin{tabular}[t]{@{}l@{}}WINE\\no gate training\end{tabular} & 78.27 & 54.13 & 13.08\\
\begin{tabular}[t]{@{}l@{}}Selector (adapted)\\with gate training\end{tabular} & 78.26--78.56 & 54.13 & 12.41--12.88\\
\bottomrule
\end{tabular*}
\end{table}

\subsubsection{Routing through answer sufficiency}
\label{sec:routing}
Table~\ref{tab:routing-rules} shows that WINE outperforms both confidence gates at 44.50\% mean 8B calls with shared answers; \cref{fig:accuracy-calls}(b) visualizes the comparison. Top-1 margin scores attribute confidence; conditional margin scores program-answer confidence. Structure-first keeps zero-read answers and uses Top-1 margin otherwise.

Relative to WINE, Top-1 margin makes 120.0 additional errors, including 111.0 on zero-read questions: it escalates despite a structure-fixed answer. Conditional margin makes 73.4 additional errors, of which 71.4 $[42.4,100.4]$ occur outside this subset. Calls are matched over the full panel.

Structure-first makes 40.8 fewer errors by escalating WINE singletons, but 27.0 more by retaining modal answers that WINE escalates. MLP makes 56.68 and 28.48 fewer errors on these respective switches. Appendix~\ref{app:routers} details the counts. With shared supervised heads, MLP adds correctness learning and attains higher holdout accuracy; WINE derives acceptance directly from the execution alternatives used for acquisition.

The supporting comparison in Table~\ref{tab:published-main} adapts Reliable VQA Selector~\citep{whitehead2022reliable} to shared heads and answers with additional correctness labels. Accuracy includes 8B fallback; ranges span three seeds. Fit 0 uses 45.87\% calls; Table~\ref{tab:routing-rules} averages five fits for MLP. WINE attains similar observed accuracy to the Selector without gate training, reusing the acquisition trace for local acceptance.

\begin{table}[!t]
\centering\small
\setlength{\tabcolsep}{3.5pt}
\caption{Population-weighted live GQA accuracy and cascade cost on 256 questions.}
\label{tab:cost-main}
\begin{tabular*}{\columnwidth}{@{\extracolsep{\fill}}lrrrr@{}}
\toprule
System & Acc. (\%) & \shortstack{Total VLM\\input tokens} & \shortstack{Mean\\latency (ms)} & \shortstack{Calculated\\model\\TFLOPs}\\
\midrule
8B alone & 77.77 & 232.88 & 181.42 & 4.134\\
WINE & 78.57 & 214.95 & 154.54 & 2.201\\
\shortstack[l]{Structure-\\first} & 80.92 & 229.45 & 162.53 & 2.215\\
\shortstack[l]{MLP-routed\\cascade} & 79.74 & 238.43 & 180.48 & 2.256\\
\bottomrule
\end{tabular*}
\end{table}

\subsubsection{Computational consequences}
\label{sec:cost}
Table~\ref{tab:cost-main} reports population-weighted accuracy and per-question cost for complete cascades. The MLP-routed cascade uses the Supervised MLP router in Table~\ref{tab:routing-rules}. VLM input tokens sum visual and text positions across small-model reads and routed 8B calls; the MLP processes numerical features and adds no tokens.

FLOPs count neural operations, including the MLP router. Appendix~\ref{app:token-accounting} reconstructs input tokens across five fits on the full holdout: 237.15 for 8B alone and 211.28 for WINE, a 10.91\% decrease.

Alternative routers attain higher observed accuracy, while WINE reaches a lower mean-cost operating point.

WINE takes 154.54\,ms versus 181.42\,ms for 8B alone: 14.81\% lower mean latency and 46.76\% fewer calculated model FLOPs. Mean latency also falls by 14.37\% $[8.94,20.07]$ versus the MLP-routed cascade and 4.92\% $[-3.95,13.09]$ versus Structure-first. P95 latency increases by 11.11\% relative to 8B alone, reflecting sequential reads before fallback. Local acceptance ends acquisition and skips 8B inference, lowering mean cascade cost under supplied structure.

\begin{table}[!t]
\centering\small
\caption{Cascade errors under supplied and predicted structural inputs.}
\label{tab:structure-main}
\setlength{\tabcolsep}{3pt}
\begin{tabular*}{\columnwidth}{@{\extracolsep{\fill}}lrrr@{}}\toprule
 & Supplied & Predicted & Predicted\\
 & $G,P$ & relations & rel.\,+\,$P$\\\midrule
Relations & Annotated & Box rules & Box rules\\
Program $P$ & Supplied & Supplied & 2B parser\\\midrule
8B calls (\%) & 44.91 & 55.80 & 70.73\\
Cascade errors & 5,497 & 6,219 & 7,888\\
\shortstack[l]{$\Delta$ errors vs.\\conditional margin} & $+181$ & $+187$ & $+51$\\
$\Delta$ errors vs. 8B & $+623$ & $-99$ & $-1{,}768$\\
\shortstack[l]{Error among zero-read\\returns (\%)} & 0.0 & 19.7 & 69.9\\\bottomrule
\end{tabular*}
\end{table}

\subsubsection{Dependence on structural reliability}
\label{sec:boundary}
Table~\ref{tab:structure-main} evaluates supplied and predicted inputs on a separate 26,018-question GQA panel with one fit. The first change replaces annotated relations with box-based spatial rules; the second also predicts programs with Qwen3-VL-2B. Object names and boxes remain supplied. Each $\Delta$ errors row names its reference. Appendix~\ref{app:structure} gives the prediction procedure.

With predicted relations, $\Delta$ errors is $+187$ $[139,237]$ versus conditional margin and $-99$ versus 8B; adding predicted programs changes the latter to $-1{,}768$. Zero-read error rises from 0\% to 19.7\% and 69.9\%. Attribute reads leave relations and program semantics fixed, so faithful inputs remain necessary for reliable local acceptance.

Controlled input corruption in Appendix~\ref{app:input-noise} tests edits to 0--50\% of eligible questions on the 10,509-question holdout, across three corruption seeds and five fixed attribute-head fits. At 20\% of eligible relation inputs, 5.06\% of all questions change and accuracy falls from 78.29\% to 77.86\%. Editing 20\% of eligible programs changes 13.25\% overall and yields 75.04\%. Wrong local / all rises from 7.38\% to 10.96\%, while 8B routing falls from 44.50\% to 43.09\%. Rates apply to eligible question inputs, with different sets for each edit. Type-valid program errors can yield wrong singletons. Agreement establishes sufficiency under the supplied semantics; semantic fidelity remains an upstream requirement.

\subsection{Analysis of answer sufficiency}
\label{sec:evidence-analysis}
We relate execution and candidate evidence to local acceptance, then examine the resulting observation depth.

\paragraph{Answer containment.}
\label{sec:execution-audit}
Local stopping relies on preserving possible answers during execution. The execution audit checks 9,477 abstract cases for 13 templates over three objects, two colors, and two materials. All 51,200 defined concrete executions, including 29,312 singleton-to-world comparisons, preserve answer containment, supporting the transfer rules in Section~\ref{sec:guarantee}. This isolates executor behavior from attribute prediction quality.

\paragraph{Candidate evidence and acceptance.}
\label{sec:candidate-breadth}
Mapped GQA annotations assess the attribute candidates supplied to the executor. At $\tau=0.8$, cell overlap is 89.67\% across 191 of 213 eligible cells; joint overlap is 82.05\% across 64 of 78 fully mapped questions. Appendix~\ref{app:structured-audits} discusses missing labels, aliases, and multiple values that restrict these rates to annotation compatibility.

The 128-question GQA development audit isolates candidate breadth by holding predictions fixed. Increasing candidate mass from $\tau=0.6$ to $0.8$ and $0.95$ reduces wrong local returns from 16 to eight and two, while increasing 8B calls from 40 to 60 and 94. Local risk falls from 18.18\% to 11.76\% and 5.88\%; final cascade correctness is 99, 102, and 101 questions. Broader candidates make answer agreement more demanding; cascade accuracy also depends on the additional 8B answers. Appendices~\ref{app:protocols} and~\ref{app:structured-audits} specify the threshold and paired intervals.

On 1,024 CLEVR questions with one SmolVLM-256M fit and no 8B fallback, increasing $\tau$ from 0.6 to 0.95 likewise lowers local coverage from 93.55\% to 85.16\% and raises local accuracy from 98.54\% to 99.77\%. Local accuracy is the complement of local risk. At $\tau=0.8$, WINE returns five wrong answers versus 17 for Strict top-1 execution. Appendix~\ref{app:structured-audits} details the retention protocols.

\paragraph{Stopping before complete observation.}
These acceptance conditions translate into variable observation depth. GQA has 1.55 reachable cells per question, with one cell in 59.83\% of questions and two in 32.68\%. WINE reads 1.26 cells and stops before pool exhaustion on 17.17\%, including 6.38\% zero-read returns. Unresolved questions use 1.55 of 1.57 available cells. Early acceptance thus coexists with nearly exhaustive reads before 8B fallback; Table~\ref{tab:read-budget} tests selection with larger CLEVR pools.

Read-all acquires candidate sets for all $4N$ cells before set-valued execution, testing whether answering requires complete observation. On the 1,024-question CLEVR panel, it yields 90.62\% accuracy with SmolVLM-256M, versus 91.41\% for WINE at the full cap. WINE checks sufficiency after each read, allowing part of the budget to remain unused. This comparison jointly changes acquisition and stopping; Table~\ref{tab:read-budget} instead holds stopping fixed to isolate read ordering. Appendix~\ref{app:read-order} details the protocols.

The controller therefore treats further observation as conditional on answer ambiguity, allowing a local return once the retained alternatives agree.

\section{Limitations and future work}
\label{sec:limitations}
Input fidelity, observation budgets, and runtime conditions define the evaluation scope. Faithful structure, program semantics, and truth-containing candidates underpin local correctness; predicted-input and controlled-edit panels examine specific error processes. Acquisition gains concern attribute-read allocation using trace ambiguity, with comparable accuracy under matched crop budgets in Appendix~\ref{app:read-order}. Runtime measurements cover one GPU with supplied structure; the Structure-first latency interval includes zero.

Future work will investigate an experience memory linking question and visual context to observation strategies and correction outcomes. Retrieved experiences would guide observation and answer revision. Reward-based updates would favor corrections and penalize changes from correct to incorrect answers, with VLM weights fixed. Each revision would be assessed against the current image. Evaluation would separate retrieval benefits from additional inference and measure correction, preservation of correct answers, and total cost on new images.

\section{Conclusion}
\label{sec:conclusion}
WINE coordinates observation, local answering, and escalation through answer sufficiency without separately training a correctness gate. On GQA, it outperforms the tested attribute- and answer-confidence gates at matched 8B usage. Attribute observations reduce 8B usage from 93.62\% for Zero-read + 8B to 44.50\% at 78.29\% accuracy. On CLEVR, intermediate-state guidance resolves more answers under half-scene attribute-read caps across backbones with shared execution and stopping. The live benchmark also shows lower mean latency and calculated model FLOPs relative to 8B alone. Shared execution thus connects further perception to local answer sufficiency.

\label{sec:main-end}

\section*{LLM Usage Statement}
OpenAI Codex was used to assist with the analysis of saved experimental results, manuscript polishing, and LaTeX preparation. It also supported the layout of figures using recorded data and user-provided icons.

{
\small
\bibliographystyle{ieeenat_fullname}
\bibliography{cvpr2027_conference}
}
\clearpage
\maketitlesupplementary
\section*{Reproducibility statement}
The supplementary material details the answer-sufficiency criterion, the acquisition rule, and the cascade controller that applies both through a shared execution state. Appendices~\ref{app:execution} and~\ref{app:protocols} specify the semantics and experimental protocols. The accompanying Overleaf package contains the manuscript sources and the three PNG figures used in the paper; it is a document source package, not an experimental artifact release. Appendix~\ref{app:routers} isolates escalation with shared answer arrays. Appendix~\ref{app:compute} analyzes structural prediction and computational costs, including live routing at matched 8B call counts. Appendix~\ref{app:structured-audits} separates candidate-set assumptions, attribute-read selection, and matched-crop diagnostics. Appendix~\ref{app:comparisons} reports published-component comparisons and controlled input corruption, including adaptation details, data splits, and error-rate denominators.

\appendix
\section{Execution semantics and conditional correctness}
\label{app:execution}

Set-valued execution provides a conditional guarantee for singleton acceptance. A world assigns one attribute value to each object--attribute cell while keeping the supplied structure fixed. The typed executor over-approximates the answers of defined concrete executions: abstract states enclose concrete object sets by $M\subseteq S\subseteq U$, references by $o\in C$, and scalar or Boolean values by $v\in V$. The transfer rules for the supported GQA operations preserve these relations.

\paragraph{Object sets and references.}
Name selection returns an exact set, $M=U$. Every evidence set is nonempty, since an unread cell holds the full attribute domain and a read returns at least the most probable value. Filtering by an allowed value set $A$ therefore gives $M'=\{o\in M:E(o,a)\subseteq A\}$ and $U'=\{o\in U:E(o,a)\cap A\neq\varnothing\}$; a positive filter uses $A=\{v\}$, for which the first condition is the $E(o,a)=\{v\}$ of Equation~\ref{eq:filter} in the main paper, and a negated filter uses the domain complement of $\{v\}$. Set-valued relations from the supplied graph map $(M,U)$ to $(R(M),R(U))$. An operation that requires a single object takes the reference $C=U$ and fails when $U$ is empty or $|M|>1$, since no consistent world then has exactly one such object.

\paragraph{Values and comparisons.}
Section~\ref{sec:execution} describes attribute-query unions and Boolean combinations. Name queries similarly return the union of the supplied names of the objects in $C$. Choices intersect query values with the supplied options and fail when the intersection is empty; relation choices retain, for each candidate object, the relation option that holds uniquely, and fail when no candidate has one. Existence retains true when $U\neq\varnothing$ and false when $M=\varnothing$. Attribute verification retains true if some candidate object may have the value and false if some candidate may have another value; negation swaps the outcomes, and relation verification uses the fixed structure. Attribute equality between two references retains true for intersecting value sets and false unless both sets are the same singleton; difference negates these outcomes.

\paragraph{Proof and scope.}
Induction over the well-typed program establishes Equation~\ref{eq:containment} in the main paper: each transfer retains every defined concrete outcome, and answer conversion preserves inclusion. If the true scene is retained, the corresponding answer therefore belongs to the final abstract set, and a singleton must equal that answer, proving Proposition~\ref{prop:sufficiency} in the main paper. This stopping result holds independently of read order. Relating the program answer to the natural-language question also requires a correct program. Ignored correlations between intermediate alternatives can leave infeasible answers, so the abstraction is sound but incomplete. Failed hypothetical executions affect selection scores without pruning actual evidence. Algorithm~\ref{alg:trace} uses the heuristic read-selection score from Equation~\ref{eq:selection} in the main paper.

\begin{algorithm*}[t]
\caption{Local GQA controller used by WINE}
\label{alg:trace}
\begin{algorithmic}[1]
\Require Image $I$, program $P$, structure $G$, attribute reader $\operatorname{Read}$, optional budget $B$
\State $E(c)\gets D_{a(c)}$ for every cell; $Q\gets\varnothing$
\While{true}
  \State Execute $(P,G,E)$ abstractly to obtain the trace and answer set $\mathcal A$
  \State \textbf{if} execution fails \textbf{then return} unresolved
  \State \textbf{if} $|\mathcal A|=1$ \textbf{then return} the sole answer in $\mathcal A$
  \State $\mathcal C_t\gets$ reachable cells available to $\operatorname{Read}$ and not in $Q$
  \State \textbf{if} $|Q|\geq B$ or $\mathcal C_t=\varnothing$ \textbf{then return} unresolved
  \State Score cells by Equation~\ref{eq:selection}; omit failed hypothetical executions
  \State Choose the first minimum-score cell $c$; if all scores fail, choose the first cell
  \State $S_c\gets\operatorname{Read}(I,G,c)$; $E(c)\gets S_c$; $Q\gets Q\cup\{c\}$
\EndWhile
\end{algorithmic}
\end{algorithm*}

The reader returns a nonempty $S_c$ by cached lookup for primary GQA accuracy or live perception for timing as described in Appendix~\ref{app:matched-runtime}. Only the selected cell is observed. The variables $E,Q$ in Algorithm~\ref{alg:trace} correspond to the current $E_t,Q_t$ in the main paper; the four stages correspond to \cref{fig:overview}.

Singleton checks precede budget checks, including after the final permitted read. Primary GQA uses $B=\infty$; ties use lexicographic object ID, then the first occurrence of the attribute in the program. A local return is final; unresolved invokes 8B on $(I,q)$ in the cascade and counts as abstention in standalone evaluation.

\paragraph{Scope of local correctness.}
Execution guarantees answer containment conditional on the supplied structure, program, and evidence. Agreement with an annotated answer checks consistency with those annotations; visual truth and program fidelity require separate validation. Appendix~\ref{app:structured-audits} reports candidate--annotation compatibility, finite implementation checks, and failures of the premises. Candidate sets use the fixed-mass construction specified in Appendix~\ref{app:protocols}; Section~\ref{sec:limitations} states the scope of the resulting guarantee.

\paragraph{Precision of the abstraction.}
For the fixed seed-0 fit, exact enumeration of the retained sets on unresolved training-split questions finds 14 singletons, 13 of them correct, in 12,073 enumerated cases; 10 of the 12,083 unresolved cases exceed the 200,000-world limit. All 740 unresolved validation-B cases are enumerated, yielding no singleton. Enumeration resolves abstraction slack within the retained candidates, preserving any omissions introduced by perception.

\section{Training and evaluation protocols}
\label{app:protocols}

Attribute-head training, optional supervised-router fitting, and cascade evaluation use distinct data partitions. An attribute-head fit contains the trained heads for all attributes; a router fit trains a separate routing model. The routing comparisons in Appendix~\ref{app:routers} follow this separation. Primary GQA accuracy evaluation reuses saved attribute and 8B predictions; live inference is evaluated separately in Appendix~\ref{app:matched-runtime}.

\paragraph{Structural input preparation.}
Primary GQA retains object IDs, names, boxes $(x,y,w,h)$, and relation lists from annotated scene graphs, removing attribute fields. A fixed parser reads the semantic operations, arguments, and dependencies annotated for each question. Argument resolution ignores annotated object-ID suffixes; graph IDs remain internal keys. Programs using same-attribute relation shortcuts are excluded. The toolbox program comprises \texttt{select(toolbox)} $\to$ \texttt{filter\_color(red)} $\to$ \texttt{query(material)}; the query implicitly requires one referent. Annotations remain available offline for head training, eligibility selection, and scoring.

CLEVR/CLOSURE supply object inventories, pixel crop centers, and spatial relations from CLEVR scene annotations. Crops are fixed $104\times104$ windows around the centers, shifted within image bounds. Dataset functional programs supply $P$; the intermediate \texttt{\_output} annotations from CLOSURE are discarded. Inference withholds size, color, material, and shape labels and instead uses crop-derived candidate sets. Both primary protocols take $G,P$ as supplied inputs.

\paragraph{GQA attribute heads.}
Supplied boxes are enlarged by 10\% on each side, with minimum intended extent 32 pixels, and clipped to the image. Last-hidden-state features of the frozen SmolVLM-256M, run in FP16, use the native chat template with ``What color is the \{name\}?'' or ``What is the \{name\} made of?'' and the corresponding answer prefix. Standardized features feed two-layer MLP heads with width 256, GELU and dropout .1, trained in FP32 for 40 epochs with cross-entropy, AdamW with learning rate $3\times10^{-4}$ and weight decay .01, batch 256 and a cosine schedule, on 25,569 color and 3,026 material labels. The 5,614 head-development images are shuffled with Python seed 0 into 4,772 fitting images, 561 selection images, and 281 separate calibration images. Alternating positions in the selection split form subset A with 281 images and subset B with 280 images. Subset A selects the epoch by top-1 accuracy and fits per-attribute temperatures and APS thresholds on temperature-scaled probabilities at $\alpha=.05$ for feature selection. Subset B compares five feature variants: pooled image-token features with or without a name embedding, name-prompted final states of SmolVLM-256M and -500M, and SigLIP features with a name embedding. Selection minimizes mean set size subject to at least 93\% pooled empirical candidate coverage, breaking ties by lower negative log-likelihood. This selects the name-prompted SmolVLM-256M state. Five attribute-head fits retain this feature variant and all image partitions, varying only initialization.

\paragraph{Diagnostic calibration.}
The separate 281-image calibration split supplies 1,389 color and 180 material labels per attribute-head fit. These images are excluded from head fitting, epoch selection, and feature selection. For each fit, saved untempered probabilities on these cells yield per-attribute APS thresholds at $\alpha=.05$ for the diagnostic singleton feature. Separately, temperatures minimize negative log-likelihood over 61 log-spaced values in $[.25,4]$ for calibrated-posterior diagnostics and router features. The diagnostic APS thresholds are computed from raw probabilities, not from temperature-scaled probabilities. These quantities are fitted on the separate calibration split rather than reused from subset A. Primary WINE uses untempered probabilities and the fixed mass threshold below; these diagnostic calibrations do not determine its candidate sets. We make no question-level conformal coverage claim.

\paragraph{Synthetic perception and partitions.}
The main CLEVR acquisition panel uses 1,024 distinct validation images, with one question sampled per image using seed 20260924 after excluding earlier diagnostic panels. Each frozen backbone has four independently fitted attribute heads for size, color, material, and shape. The SmolVLM backbones run in FP16 and Qwen3-VL-2B in BF16; these acquisition comparisons use no QA adapter. A constant prompt, ``Describe this object.'', accompanies each $104\times104$ crop. Features average final-layer image-token states. SmolVLM disables image splitting and uses a processor longest edge of 512; Qwen uses the native processor. The 2,500 CLEVR training images are partitioned into 2,250 fitting and 250 development images using Python shuffle seed 0. The resulting 14,622 fitting crops provide 58,488 attribute labels; 1,648 development crops select the heads. Training standardizes features using fitting data and uses width-256 MLP heads, GELU, dropout .1, cross-entropy, AdamW with learning rate $3\times10^{-4}$ and weight decay .01, batch 256, and a 40-epoch cosine schedule. Each head uses initialization seed 0 and the epoch of highest development accuracy. No acquisition policy is trained.

CLOSURE contains 128 questions from each of seven compositional families, totaling 896 distinct validation images, excluding earlier CLOSURE diagnostics. It uses an earlier CLEVR-trained SmolVLM-256M attribute-head checkpoint set, fitted on 2,250 images with 14,688 crops and selected on 250 development images; it differs from the CLEVR-panel checkpoint set above. The architecture and optimization are the same, but the image split uses NumPy seed 0. No parameters are fitted on CLOSURE. Functional programs are supplied, so this comparison concerns execution and acquisition under new compositions rather than language-to-program generalization.

\paragraph{Observation reuse and tie order.}
A synthetic crop produces all four attribute distributions in one backbone call. Cell-budget inference reveals only the requested attribute and caches the remaining distributions within the question; a later request for the same object needs no new encoding. Consequently, equal cell budgets need not imply equal crop counts. Synthetic rankings compute the reduction from the current ambiguity using exact floating-point comparisons and retain the first maximizing cell in the executor pool. The GQA ranking instead compares residual ambiguity with tolerance $10^{-12}$. These implementation-specific tie rules are fixed across the corresponding WINE and Final-set controls.

\paragraph{Mass threshold.}
\label{app:mass-threshold}
Primary inference uses untempered head probabilities with $\tau=.8$, fixed across the reported GQA evaluations. A saved CLEVR development analysis evaluates $\{.6,.7,.8,.9,.95\}$ on 1,648 crops comprising 6,592 attribute cells across five attribute-head fits. Candidate coverage is the fraction of labeled cells whose candidate set contains the labeled value. The smallest value attaining 99\% mean cell-level candidate coverage is $.8$, with 99.12\% candidate coverage; the same value satisfies the criterion for each fit. This provides an empirical rationale for the operating point, not a question-level coverage guarantee. Calibration-crop temperatures and APS thresholds serve only diagnostic controls and router features.

\paragraph{Eligibility and panel separation.}
Eligible questions use supported color/material programs and reproduce the annotated answer on the annotated graph. This retains 20,258 of 132,062 balanced-validation questions, or 15.34\%; 95,612 of the 111,804 exclusions read no attribute. The primary train-split router pool contains 26,223 questions on 9,345 images, split by image into 15,714 fitting questions on 5,607 images and 10,509 holdout questions on 3,738. Validation A has 1,024 questions on 1,024 images and validation B 1,630 on 577. The entire head-development pool is disjoint from routing panels.

The auxiliary panels serve separate purposes. The compute build has 2,005 questions on 713 images and overlaps validation A and B by 254 and 980 questions. The 26,018-question structure study uses an earlier version of the program parser. These are selected symbolic-reasoning panels, and the train-split holdout is not claimed to be unseen by the external model. The public pretraining mixture of SmolVLM also includes GQA through The Cauldron~\begin{NoHyper}\citep{laurencon2024idefics2}\end{NoHyper}.

\paragraph{Router features and fitting.}
Correctness targets compare shared local predictions with existing GQA answers; these existing annotations provide all fitting targets, independently of 8B outcomes. Answer features comprise 18 quantities: posterior scores and defined fractions at both sample counts, head margins and confidence, cell/object counts, top-1 definedness and question type. Six additional executor features encode WINE and diagnostic crop-calibrated singleton status, zero-read status, reads, exact-answer-set size and enumeration availability. Five router seeds accompany each attribute-head fit. An image-disjoint 75/25 split selects hyperparameters by AUROC before refitting without early stopping. Logistic regression (LR) searches $C\in\{.05,.3,1,5\}$; histogram gradient boosting (HGB) searches learning rates $\{.05,.1\}$, leaves $\{15,31\}$ and minimum leaf sizes $\{20,50\}$; the multilayer perceptron (MLP) searches layers $(32),(64,32),(128,64)$ and regularization $\{10^{-4},10^{-2}\}$. Supervised MLP denotes the MLP router with both answer and executor features; the answer-feature-only ablation is labeled separately. Holdout routers use the fitting partition; validation routers use the full training panel with internal selection.

\paragraph{External answers and scoring.}
Qwen3-VL-8B-Instruct uses BF16, the native chat template and the instruction ``Answer with only the short answer, without explanation.'' Greedy decoding allows at most 16 new tokens. Exact matching normalizes case, punctuation, articles and number words. Every routing rule reuses the same saved predictions. Accuracy includes unresolved outputs as errors.

\paragraph{Descriptive decomposition.}
Local-return and escalation groups, group accuracies, and read counts are computed from saved per-question local and 8B correctness and reachable-cell counts; no model, head or router is rerun. Group accuracies divide five-fit mean correct counts by five-fit mean group sizes, and the resulting error counts reproduce the published totals of 2,281.0, 2,475.0 and 2,987.8 exactly.

\section{Complete GQA routing comparisons}
\label{app:routers}

The complete cascade changes both local answering and escalation relative to 8B alone. Fixing both answer arrays isolates the routing contribution by varying only the questions assigned to 8B. A separate independent-answerer comparison evaluates changes to local answering.


\paragraph{Shared-answer routing comparison.}
Table~\ref{tab:app-gqa-free} lists the gates that use no answer-correctness labels on all three panels. Entries are error counts averaged across five fits, producing fractional means. The local-only and 8B-alone rows provide no-call and all-call references. Each routing row shares a local-answer array containing the WINE singleton if available and otherwise the modal result of 256 sampled executions, together with the saved larger-model answers. WINE escalates every unresolved case; the modal local candidate permits alternative gates to retain these cases locally. Alternatives make exactly as many calls as WINE within each attribute-head fit and panel: mean call counts and rates are 4,676.4/10,509 at 44.50\%, 471.6/1,024 at 46.05\%, and 731.6/1,630 at 44.88\%. The corresponding counts range over 4,580--4,820, 457--491 and 720--747 across fits. Accuracy differences therefore measure which questions are escalated, with both potential answers held fixed.

\begin{table*}[!t]
\centering
\caption{Errors for gates without answer-correctness labels on three GQA panels (five-fit means).}
\label{tab:app-gqa-free}
\footnotesize
\setlength{\tabcolsep}{7pt}
\begin{tabular}{@{}lrrr@{}}
\toprule
Gate & Holdout & Validation A & Validation B \\
\midrule
\rowcolor{traceRow}
\textbf{WINE} & 2,281.0 & 243.0 & 397.6 \\
Structure-first ($S_0+T$) & 2,267.2 & 242.2 & 401.0 \\
Top-1 margin & 2,401.0 & 251.2 & 424.0 \\
Posterior confidence & 2,508.4 & 262.8 & 432.8 \\
Crop-calibrated confidence & 2,572.4 & 267.8 & 443.4 \\
Conditional share, $K=256$ & 2,383.2 & 253.4 & 411.6 \\
Conditional margin, $K=256$ & 2,357.2 & 251.6 & 412.6 \\
Conditional share, $K=1{,}024$ & 2,383.6 & 252.0 & 413.8 \\
Conditional margin, $K=1{,}024$ & 2,354.4 & 250.8 & 413.4 \\
\midrule
Local only (no calls) & 2,987.8 & 309.2 & 511.6 \\
8B alone (all calls) & 2,475.0 & 254.0 & 413.0 \\
\bottomrule
\end{tabular}
\end{table*}

\paragraph{Gate scores.}
Top-1 margin takes the minimum gap between the two highest attribute probabilities over cells pairing structurally reachable objects with attribute types used in the program, including unread cells. Executing the most likely attribute values partitions questions into defined and undefined top-1 executions. Undefined executions are escalated first, with each group ordered by increasing margin. Posterior gates sample complete scenes from the head distributions and execute the program \(K\) times. With \(d\) defined executions and the two largest defined-answer counts \(a_1\) and \(a_2\), posterior confidence is \(a_1/K\), conditional share is \(a_1/d\), and conditional margin is \((a_1-a_2)/d\) for \(K\in\{256,1{,}024\}\). Structure-first ($S_0+T$) keeps answers already determined before any read and otherwise follows top-1 margin. These controls distinguish confidence in attributes, confidence in program answers, and preservation of structure-fixed answers.

\paragraph{Validation routing outcomes.}
Validation accuracy is decomposed into local-return and escalated subsets. On validation A, WINE keeps 53.95\% of questions, where local answers reach 86.75\% against 84.76\% for 8B, and escalates the rest, where the modal local candidate reaches 49.96\% against 63.99\%. On validation B the corresponding values are 55.12\%, 83.37\% against 81.66\%, and 50.49\% against 66.07\%. After-read local and 8B accuracies are 85.74\% versus 85.94\% on validation A and 81.29\% versus 82.74\% on validation B, similar to the holdout. The 39 and 100 zero-read questions are answered without error locally.

\paragraph{Sampling details.}
Posterior confidence, conditional share, and conditional margin are assigned zero when no sampled execution is defined. The 256 draws form a prefix of the 1,024 draws with a fixed cell order. Crop-calibrated scores use temperatures fitted on calibration crops. All gates use supervised attribute heads even when the routing scores require no answer-correctness labels.

\paragraph{Error exchanges between routing rules.}
The zero-read subsets contain 670, 39 and 100 questions per fit, with no WINE, $S_0+T$, or MLP errors. Holdout differences therefore arise among the other 9,839 questions. A routing switch either sends a WINE singleton to 8B or accepts a modal local answer that WINE escalates. In each case, corrected errors are offset by newly introduced errors.

For Structure-first, switching singletons to 8B corrects 153.6 errors and introduces 112.8, yielding 40.8 fewer errors. Accepting modal answers corrects 170.8 errors but introduces 197.8, yielding 27.0 more errors. These opposing effects give $\Delta$ errors of $-13.8$ relative to WINE.

MLP improves both switch types. Sending singletons to 8B corrects 121.72 errors and introduces 65.04; accepting modal answers corrects 92.60 and introduces 64.12. The resulting decreases are 56.68 and 28.48 errors, totaling 85.16. Counts average five attribute-head fits for $S_0+T$ and 25 attribute-head--router combinations for MLP. Section~\ref{sec:routing} relates this routing trade-off to the shared-trace design.

\paragraph{Comparator selection.}
The strongest of six posterior scores is selected on each panel: conditional margin at $K=1{,}024$ on the holdout and validation A, and conditional share at $K=256$ on validation B. The corresponding $\Delta$ errors values are 73.4 $[44.2,102.8]$, 7.8 $[0.4,15.6]$ and 14.0 $[0.6,27.6]$. The validation intervals barely exclude zero and are pointwise, without adjustment for selecting the comparator.

\subsection{Supervised routers and variation across fits}

Table~\ref{tab:app-gqa-learned} compares supervised routers and random/oracle references using the same local and 8B answers and the WINE call count within each panel. Within each router family, the answer-only and added-executor conditions isolate the contribution of execution features. Appendix~\ref{app:protocols} defines the features and fitting procedure.

\begin{table*}[!t]
\centering
\caption{Mean errors for supervised routers and reference policies at matched 8B call counts.}
\label{tab:app-gqa-learned}
\footnotesize
\setlength{\tabcolsep}{7pt}
\begin{tabular}{@{}lrrr@{}}
\toprule
Gate & Holdout & Validation A & Validation B \\
\midrule
LR, answer features & 2,285.4 & 243.4 & 398.0 \\
LR, $+$ executor features & 2,235.2 & 240.6 & 393.2 \\
HGB, answer features & 2,224.8 & 238.2 & 394.0 \\
HGB, $+$ executor features & 2,211.4 & 235.8 & 391.2 \\
MLP, answer features only & 2,208.8 & 235.4 & 392.2 \\
Supervised MLP & 2,195.8 & 234.3 & 389.5 \\
\midrule
Random & 2,759.5 & 283.8 & 467.3 \\
Oracle & 1,402.6 & 150.4 & 251.4 \\
\bottomrule
\end{tabular}
\end{table*}

\begin{table*}[!t]
\centering
\caption{Holdout error ranges and paired differences from WINE with 95\% CIs.}
\label{tab:app-gqa-intervals}
\footnotesize
\setlength{\tabcolsep}{5pt}
\begin{tabular}{@{}lrr@{}}
\toprule
Comparator & Errors, run range & $\Delta$ errors \\
\midrule
WINE & 2,267--2,293 & --- \\
Structure-first ($S_0+T$) & 2,253--2,283 & $-13.8\;[-63.6,34.4]$ \\
Top-1 margin & 2,396--2,403 & $120.0\;[66.0,172.0]$ \\
Conditional margin, $K=1{,}024$ & 2,340--2,365 & $73.4\;[44.2,102.8]$ \\
Supervised MLP & 2,170--2,224 & $-85.2\;[-117.6,-53.8]$ \\
8B alone & 2,475--2,475 & $194.0\;[132.0,254.2]$ \\
\bottomrule
\end{tabular}
\end{table*}

\paragraph{Sources of variation.}
The run ranges in Table~\ref{tab:app-gqa-intervals} span five attribute-head fits, or 25 attribute-head--router runs for MLP. $\Delta$ errors is comparator errors minus WINE errors; positive values favor WINE. The bracketed paired 95\% CIs quantify uncertainty in this difference, not the range across runs.

The holdout error count of WINE has standard deviation 9.35 across attribute-head fits. The invariant, error-free zero-read subset contributes none of this variation. Router-seed variation occurs only for the MLP; the within-fit holdout standard deviation averages 12.3 across attribute-head fits. Using Supervised MLP as reference, $\Delta$ errors is $-85.2$ $[-117.6,-53.8]$ on the holdout, $-8.7$ $[-17.8,0.2]$ on validation A, and $-8.1$ $[-20.8,4.0]$ on validation B. The MLP is therefore the strongest holdout comparator. Both methods use supervised attribute heads. The MLP gains accuracy by learning when to override the shared local answer, whereas WINE uses the execution trace for both acquisition and acceptance without fitting a correctness gate. This comparison measures the routing trade-off for a common answerer.


\paragraph{Decomposition by evidence requirement.}
The holdout contains 670 zero-read returns and 9,839 other questions. This partition records initial acceptance by WINE. Relative to 8B
alone, WINE has $\Delta$ errors of $+174.0$ $[147.0,202.0]$ and $+20.0$ $[-34.8,75.8]$ in these subsets. Relative to conditional margin with $K=1{,}024$, the values are $+2.0$ $[0.2,4.8]$ and $+71.4$ $[42.4,100.4]$, respectively. These GQA routing decompositions reuse the five primary attribute-head fits, with
paired 95\% intervals. Call matching applies to the full panel; subsets inherit the full-panel
policies without reranking or rematching. Even with zero-read outcomes fixed across fits, image-cluster resampling contributes uncertainty.

\paragraph{Uncertainty estimation.}
Paired percentile intervals use 10,000 bootstrap draws that cross attribute-head fits and image clusters, using sampled question counts as rate denominators and resampling router seeds within each drawn fit for learned gates. The bootstrap resamples fixed routing outcomes; calls are matched on each original fit and panel. Intervals are pointwise and unadjusted across the 15 principal gates and three panels. Intervals spanning zero admit either ordering of the compared methods; equivalence would require a prespecified tolerance.

\paragraph{Identifiability from saved predictions.}
The saved ranking inputs permit reconstruction of $S_0+T$. MLP scores and call masks were not retained, but per-question errors identify the routing choice wherever local and 8B correctness differ. Choices with equal correctness remain unidentified and contribute zero error difference. All reported error exchanges are therefore identifiable; these are descriptive counts, with no new training or confidence intervals.

\subsection{Independent local answering}
\label{app:independent-answerer}
The independent-answerer comparison evaluates changes to local prediction in addition to routing. This historical analysis uses the 10,509-question holdout, executes the top-1 attribute assignment concretely, escalates undefined executions first, and ranks the remaining questions. It matches the larger-model call count within each of the five attribute-head fits; Table~\ref{tab:independent-answerer} reports five-fit means with 4,676.4 larger-model calls. The independent-answerer condition changes both the local answerer and the compatible routing rule, precluding interpretation as a gate-only ablation. Undefined executions remain unavailable as local answers.

\begin{table*}[!t]
\caption{Independent-local-answer comparison on the 10,509-question GQA holdout.}
\label{tab:independent-answerer}
\centering\small
\begin{tabular}{@{}lrr@{}}\toprule
Local answerer and gate & Acc. (\%) & Errors\\\midrule
WINE singleton gate & 78.29 & 2,281.0\\
Shared answerer, $S_0+T$ & 78.43 & 2,267.2\\
Independent top-1, corrected $S_0+T$ & 77.81 & 2,332.2\\
\bottomrule\end{tabular}

\end{table*}

The corrected independent control makes 51.2 more errors than WINE on average. Table~\ref{tab:independent-decomposition} reports five-fit mean errors in subsets defined by zero-read behavior under WINE. $\Delta$ errors is top-1 errors minus WINE errors; positive values favor WINE. The advantage is concentrated on the zero-read subset; WINE makes 10.6 more errors on the remaining questions. Subsets inherit whole-panel call matching, so their error differences describe the joint changes to local answering and routing within the original allocation.

\begin{table*}[!t]
\caption{Error decomposition of the independent top-1 comparison (five-fit means).}
\label{tab:independent-decomposition}
\centering\small
\setlength{\tabcolsep}{4pt}
\begin{tabular}{@{}lrrrr@{}}\toprule
Subset & Count & WINE & Top-1 & $\Delta$ errors\\\midrule
All & 10,509 & 2,281.0 & 2,332.2 & $+51.2$\\
Zero-read & 670 & 0.0 & 61.8 & $+61.8$\\
Other questions & 9,839 & 2,281.0 & 2,270.4 & $-10.6$\\
\bottomrule\end{tabular}

\end{table*}

This historical development analysis requires the full perception probabilities and reports descriptive counts without a fresh evaluation panel or confidence intervals. Matching larger-model calls fixes one component of computation; local processing costs remain method-dependent. The independent-answerer comparison therefore characterizes the joint answerer--router system, while the shared-answer comparison isolates routing, and the two are interpreted separately.

\section{Structural inputs and computation}
\label{app:compute}

The structural-input analysis and computational evaluations use separate panels. Computational evidence comprises historical operation counts, reconstructed token volume, and paired live timing, with distinct denominators and measurement boundaries.

\subsection{Structural input prediction}
\label{app:structure}
\paragraph{Construction of predicted structure.}
Rung 1 replaces annotated relation edges with 2D band rules over supplied box centers, preserving names, boxes and programs. Rung 2 additionally predicts programs with few-shot Qwen3-VL-2B and rejects malformed programs through strict validation. Inexpressible questions remain in the panel as abstentions; they escalate first or count as errors. The study uses 26,018 train-split questions, one attribute-head fit and image-cluster intervals, separately from the five primary routing fits. The logistic router is fit on validation A and scored out of sample here; this router is distinct from the holdout routers. Boxes and names remain supplied at both rungs.

\paragraph{Effects of structural errors.}
Band edges recall 93.4\% of annotated spatial edges, but low precision creates oversized relation sets. Restricting Rung 1 to the 22,881 expressible questions still yields 5,169 cascade errors versus 5,070 for the larger model. Program predictions are exact on 28.1\% and executable on 57.1\% of the full panel. Separate pilots find 35.5\% exact naming for 155 supplied objects, 29.6\% box recall at IoU .5, 15.5\% recall with the correct name, and all referenced objects recovered on only 10\% of questions. These observations motivated keeping boxes and names supplied. These failures can invalidate the structural premises of singleton acceptance.

\subsection{Historical local work and component timings}
Table~\ref{tab:app-compute} reports mean local work per question on the separate compute panel. ``Crops'' counts unique objects and ``Reads'' counts object--attribute pairs. ``Executions'' counts abstract executions, with concrete program executions excluded from this column. WINE trades fewer reads for repeated abstract execution; these counts describe local work, while total inference time is measured separately.

Historical CPU microbenchmarks give 64.9\,$\mu$s per attribute-head evaluation, 14.3\,$\mu$s per GQA abstract execution, and 7.2\,$\mu$s per concrete execution. The head measurement uses CLEVR heads; the 12.6\,$\mu$s posterior cost includes sampling as well as concrete execution. A separate RTX 3090 run gives 0.185\,s per Qwen3-VL-8B generation, excluding loading, preprocessing, transfers and text decoding; allocated GPU memory peaks at 17.2\,GiB overall and 16.3\,GiB during loading.

The compute panel has one fit and 2,005 questions. A unique crop can support multiple attributes, whereas the evaluated GQA features require one forward pass per object--category pair. Appendix~\ref{app:structured-audits} reports an eight-image pilot measuring this small-model component separately. The component timings have different boundaries, so the current cascade comparison uses the common live boundary in Appendix~\ref{app:matched-runtime}. The $K=1{,}024$ row has no saved predictions or accuracy claim.

\begin{table}[!t]
\centering
\caption{Mean local work per question on the 2,005-question GQA compute panel.}
\label{tab:app-compute}
\footnotesize
\setlength{\tabcolsep}{3pt}
\begin{tabular}{@{}lrrr@{}}
\toprule
Method & Crops & Reads & Executions \\
\midrule
Top-1 execution & 1.55 & 1.60 & 0.0 \\
All-cell set-valued execution & 1.55 & 1.60 & 1.0 \\
WINE & 1.25 & 1.29 & 57.5 \\
Posterior, $K=1{,}024$ & 1.55 & 1.60 & 0.0 \\
\bottomrule
\end{tabular}
\end{table}

\subsection{Retrospective token accounting}
\label{app:token-accounting}

\begin{table*}[!t]
\centering\small
\caption{Reconstructed VLM input tokens per question on the GQA holdout (five-fit means).}
\label{tab:app-tokens}
\setlength{\tabcolsep}{5pt}
\begin{tabular}{@{}lrrrrrr@{}}
\toprule
 & \multicolumn{2}{c}{Small-model input} & \multicolumn{2}{c}{8B input} & Total VLM & Saved (\%)\\
\cmidrule(lr){2-3}\cmidrule(lr){4-5}
System & Visual & Text & Visual & Text & input tokens & \\
\midrule
8B alone & 0.00 & 0.00 & 207.74 & 29.41 & 237.15 & ---\\
WINE + 8B & 80.86 & 26.17 & 91.18 & 13.07 & 211.28 & 10.91\\
\bottomrule
\end{tabular}
\end{table*}

The token reconstruction estimates the model input required to acquire the recorded evidence on demand. It uses the same 10,509-question holdout and five attribute-head fits. CPU replay recovers all 52,545 read sequences and exactly matches the saved answers, stopping states, and read counts. Each selected object--attribute pair is charged one small-model forward pass; an unresolved question also receives the original image--question 8B prompt. Questions are accounted for independently, without cross-question cache reuse. These are on-demand costs for the recorded decisions; the original accuracy experiments precomputed the predictions.

We use the original prompts and image dimensions with locally retained SmolVLM processor assets and hash-verified official Qwen3-VL-8B processor files. The reconstructed 8B counts reproduce historical totals of 23,561 and 11,391 prompt tokens on the available 100- and 50-question validation checks. This reconstruction repeats no neural inference. The original processor revision was not logged, so these are reconstructed counts rather than historical per-call measurements.

Table~\ref{tab:app-tokens} reports five-fit mean total VLM input tokens per question relative to one full-image 8B call. ``Visual'' counts image positions and ``Text'' includes prompt and special tokens; both are included in the sum over small-model and routed 8B calls. WINE reduces routed 8B input by 56.04\%, while including small-model reads yields a 10.91\% total reduction. Totals and savings use unrounded counts.

Table~\ref{tab:app-tokens} excludes generated tokens. The small model generates none: heads and the executor produce the local answer. Historical 8B token IDs were not saved. Bounding each nonempty output by the recorded 1--16-token generation limit yields a total-token reduction of 10.84--14.03\%, conditional on the reconstructed inputs. This accounting bound shares the saved output length for each question across all attribute-head fits and both systems; it describes admissible token counts rather than sampling uncertainty. Computation per token depends on the model. CPU execution, head arithmetic, training, global cache construction, and prediction of the supplied structural inputs are excluded. Model-operation FLOPs and elapsed time are evaluated separately in Appendices~\ref{app:runtime-flops} and~\ref{app:matched-runtime}.

\subsection{Earlier two-system timing and FLOP convention}
\label{app:runtime-flops}

\begin{table*}[!t]
\centering\small
\caption{Earlier two-system run on 256 GQA questions. Population-weighted latency, total VLM input tokens per question, and model-operation FLOPs.}
\label{tab:app-runtime}
\setlength{\tabcolsep}{7pt}
\begin{tabular}{@{}lrrrrr r@{}}
\toprule
System & \shortstack{Total VLM\\input tokens} & \shortstack{Mean latency\\(ms)} & \shortstack{P90 latency\\(ms)} & \shortstack{P95 latency\\(ms)} & \shortstack{Calculated model\\TFLOPs} & 8B calls (\%)\\
\midrule
8B alone & 232.88 & 188.84 & 212.48 & 297.80 & 4.134 & 100.00\\
WINE + 8B & 214.95 & 160.96 & 287.61 & 320.63 & 2.201 & 45.87\\
\bottomrule
\end{tabular}
\end{table*}

\paragraph{Comparison and timing boundary.}
This earlier two-system run is reported separately from the four-system results in Table~\ref{tab:cost-main} and Appendix~\ref{app:matched-runtime}; its timings are not pooled with that comparison. We compare one full-image Qwen3-VL-8B call with live WINE on a fixed, proportionally stratified sample of 256 holdout questions on 246 images, using attribute-head fit 0 and sampling seed 20261110. The three historical strata contain 16 zero-read, 122 after-read local, and 118 escalated questions. They determine sampling weights. Newly computed SmolVLM features, trained-head probabilities, candidate sets, and the original executor determine each next read, return, or 8B call. No cross-question feature or answer cache is used.

Both systems run at batch size one on the same RTX 3090, with both backbones resident; SmolVLM uses FP16, the 8B model BF16, and CPU heads FP32 with two CPU threads. Warm-up uses one paired question from each stratum, and system order alternates across questions and repetitions. The timer runs from independently opening and decoding the image to the final answer, including crops, preprocessing, transfers, every neural call, CPU heads, candidate construction, selection, and execution. Image decoding is included even for zero-read questions. Lightweight token/trace diagnostics are included. Structure is supplied; structure prediction, training, network delay, and initial model loading are excluded. This measures a local warm deployment, not remote-server latency or local-memory savings.

\paragraph{Completed measurements and protocol deviation.}
Of three planned timing passes, two completed before an unrelated training job began sharing the GPU during the third. Analysis retains these two complete passes, totaling 512 timings per system, and excludes the third pass and incomplete synchronized component pass in full. All raw records, the original plan, and the deviation are preserved; no questions or individual slow observations were removed. The separate GPU-operator profile was not completed. FLOPs are derived from the architecture and actual recorded paths.

Table~\ref{tab:app-runtime} uses population-stratum weights and these two repetitions. Repeated measurements are averaged within question for the mean estimator; P90/P95 are weighted percentiles over all 512 observed timings per system. Total VLM input tokens include visual and text positions across all small-model reads and routed 8B calls per question; generated tokens are excluded. Calculated FLOPs include both model stages and actual generated-token lengths under the model-operation convention. This sample is distinct from the five-fit, full-holdout reconstruction in Table~\ref{tab:app-tokens}.

\paragraph{Cost analysis.}
Mean latency falls by 14.76\%, with a paired 95\% interval of 11.38--18.10\%, corresponding to $1.17\times$ speedup, whereas dense-reference FLOPs fall by 46.76\%. Input tokens fall by 7.70\% on this sample; input plus generated tokens fall by 8.10\%. These metrics characterize distinct aspects of inference cost. Local-after-read questions take 70.94\,ms on WINE versus 186.14\,ms on 8B alone. Escalated questions instead take 276.05 versus 192.22\,ms: local acquisition precedes the same 8B call. Sequential local processing and fallback therefore produce a slower escalation path alongside the lower overall mean. Neural computation under WINE comprises 0.3035\,TFLOPs for small-model reads and 1.8975\,TFLOPs for routed 8B calls per question.

The interval uses 10,000 paired image-cluster exponential-multiplier draws, sharing the weight assigned to each image across both systems and all questions for that image, then renormalizing within the three strata. It describes conditional sample uncertainty, not variation across attribute-head fits, devices, or temporal load. Weighted sample accuracy is 77.77\% for 8B alone and 78.57\% for WINE; the paired difference interval is $[-2.75,4.31]$ points. These are descriptive sample results, not replacements for the primary five-fit accuracy panel. All 256 read sequences and local endpoints match the archive in each pass; 326 of 327 acquired candidate sets match. Outputs and paths are stable across the two passes, and all 236 paired fallback generations have identical token IDs between systems.

\paragraph{FLOP definition and reconstruction.}
We count a multiply-add as two operations and include vision patch convolution, attention and MLP projections, the visual connector, decoder prefill and cached decoding, vocabulary projections, and the MLP attribute heads. SmolVLM processes 1,024 vision patches per read, merged into 64 language-model visual positions. The original SmolVLM feature-forward call also computes unused vocabulary logits at every input position; these logits are included in the count. The main visual merger in Qwen and all three deep-stack mergers are included. Generating $T$ token IDs, including EOS, requires one prefill and $T-1$ decode forwards; new runs save these IDs directly. Dense attention counts all query--key pairs. Using triangular causal prefill instead changes the sample reduction from 46.76\% to 46.74\%.

These are dominant matrix/convolution operation counts, checked against model configuration and implementation dimensions. They exclude normalization, nonlinear and other elementwise operations, embedding lookup, data movement, preprocessing, and symbolic CPU execution; the measured time includes these costs. Thus the FLOP and timing columns describe different aspects of the same live paths.

For the original 10,509-question holdout and five attribute-head fits, the same convention gives 4.203--4.433\,TFLOPs per question for 8B alone and 2.146--2.248 for WINE as the common output length varies from one to 16 tokens. Allowing unequal unknown output lengths yields a conservative 47.79--50.45\% reduction range. This conservative accounting range relaxes shared-output constraints across fits and describes on-demand neural costs for the original cached paths; it quantifies admissible costs rather than sampling uncertainty.

The environment uses an Intel i9-12900K, Windows 11, PyTorch 2.7.0+cu118, Transformers 5.8.1, and driver 595.95. Both models were resident in both arms, with 16.85\,GiB allocated for weights. Loading in the timing run took 35.43\,s for 8B and 1.24\,s for SmolVLM, outside the per-question timer. OS file caching, clocks, and ordinary desktop activity were not controlled; the separate GPU training job is handled by excluding the entire affected pass. These measurements characterize the recorded local deployment under the stated timing conditions.

\subsection{Live routing at matched 8B call counts}
\label{app:matched-runtime}

\begin{table*}[!t]
\centering\small
\caption{Four-system live comparison on 256 questions. Calls and correct answers are raw counts per pass; accuracy and costs are population-weighted.}
\label{tab:matched-runtime-full}
\begin{tabular}{lrrrrrrr}
\toprule
System & 8B calls & Correct & Acc. (\%) & \shortstack{Total VLM\\input tokens} & \shortstack{Mean latency\\(ms)} & \shortstack{P95 latency\\(ms)} & \shortstack{Calculated model\\TFLOPs}\\
\midrule
8B alone & 256 & 199 & 77.77 & 232.88 & 181.42 & 288.75 & 4.134\\
WINE & 118 & 201 & 78.57 & 214.95 & 154.54 & 320.84 & 2.201\\
Structure-first & 118 & 207 & 80.92 & 229.45 & 162.53 & 328.68 & 2.215\\
MLP-routed cascade & 118 & 204 & 79.74 & 238.43 & 180.48 & 333.79 & 2.256\\
\bottomrule
\end{tabular}
\end{table*}

The additional four-system measurement uses the same 256 questions and fixed stratum weights as Appendix~\ref{app:runtime-flops}, with attribute-head fit 0 and two new complete repetitions. It compares 8B alone with the WINE, Structure-first, and MLP-routed cascades. Both backbones remain resident, batch size is one, and the image is independently decoded for each system. Warm-up includes all four systems on one question per stratum. System order cycles across questions and reverses in the second repetition. Loading, structure construction, and training are excluded; all online perception, routing features, and fallback generation are included. Each timed path uses fresh model outputs; archived predictions serve only for consistency checks.

MLP-routed cascade denotes the complete system controlled by the Supervised MLP router in Table~\ref{tab:routing-rules}. Total VLM input tokens sum every small-VLM feature-extraction input and the 8B input when the question is routed. Counts include visual positions, prompt text, and special tokens, with generated output tokens excluded. The router receives numerical features and generates no tokens; its arithmetic contributes to measured latency and model FLOPs. Symbolic executions used to construct routing features add execution time without adding VLM tokens.

The MLP router uses the original 15,714-question, 5,607-image fitting split, hidden width 32, regularization $10^{-4}$, and seed 20260916. The 10,509-question error vector exactly reproduces the archived result before benchmarking. Each local router makes 118/256 calls per pass. Alternative masks rank feature scores on the timing panel at the WINE count, without using held-out correctness. Fresh rankings reproduce the finalized masks, and all shared local answers match the archive. The generated 8B token sequences also agree across systems and repetitions. Panel-level matching is descriptive, not a threshold selected for unseen traffic.

\paragraph{Live-score protocol amendment.}
The initial two passes used masks from archived scores. The prespecified check found that small cached-versus-live probability differences exchanged one retained and one escalated question for MLP, identically in both passes; Structure-first masks, WINE endpoints, and all shared local candidates were unchanged. We retained the entire first run as a preflight record, froze masks from the repeat-identical live scores without consulting correctness, and repeated the full four-system benchmark twice. Only the two new complete passes enter Table~\ref{tab:matched-runtime-full}. No questions, individual timings, model weights, fitting settings, or call counts were selected or changed using outcomes.

Calls and correct answers in Table~\ref{tab:matched-runtime-full} are unweighted counts per pass; accuracy and cost entries apply the fixed population-stratum weights. Repeat predictions agree. Latencies include within-question feature reuse. WINE reads only selected cells. Structure-first checks zero-read answers, computes margins over all required cells otherwise, and computes the shared local candidate only for retained questions. The MLP-routed cascade computes every original answer/execution feature, including $K=1,024$ sampled executions, calibrated posterior features, the diagnostic singleton, and bounded exact enumeration. The costs cover each complete cascade, including feature extraction, routing, and any 8B fallback. A simpler learned feature set could have a different trade-off.

For paired uncertainty, the two repetitions are averaged within question and 246 image clusters are resampled jointly 10,000 times, using seed 20261110 and fixed stratum weights. The 512 timings per system are not treated as 512 independent questions. Intervals condition on the sampled panel, one attribute-head fit, one router fit, and fixed masks; they exclude retraining and deployment variability. P95 is the inverse weighted empirical CDF of the 512 individual timings, rather than the quantile of per-question averages.

WINE reduces mean latency relative to the MLP-routed cascade by 14.37\% [8.94,20.07]\%, with an accuracy difference of -1.17 points [-3.17,0.78]. Relative to 8B, the mean-latency reduction is 14.81\% [6.88,22.86]\%. These conditional intervals characterize the sampled operating points. The Structure-first and MLP-routed cascades attain higher observed accuracy, while WINE has lower observed mean cost. Matching 8B calls leaves local feature and execution costs method-dependent; the five-fit accuracy analysis remains the primary routing comparison.

FLOPs follow the dominant matrix-operation convention in Appendix~\ref{app:runtime-flops}, using actual read receipts, image grids, and generated lengths. The MLP router adds 1,600 dense multiply-add FLOPs per question for the $24\to32\to1$ layers. Symbolic execution, random sampling, and preprocessing are not represented in model-operation FLOPs and remain included in measured latency. Supplied structure costs are excluded in every local system. The historical two-system timing above is retained as a separate run and is not pooled with these new repetitions.

\section{Candidate sets, execution, and cost audits}
\label{app:structured-audits}

These diagnostics examine candidate size, read order, and the premises of singleton acceptance using retained predictions of structured WINE. The GQA candidate-set panel contains 128 questions on 128 images and one attribute-head fit. It was previously scored and serves as development data. Thresholds and order controls are descriptive conditions; no threshold or policy is selected from the resulting scores. Large-model answers and attribute predictions are cached. The synthetic risk and crop-cost analyses use separate panels and fit counts.

\subsection{Candidate size and local answer risk}
Table~\ref{tab:gqa-tau-audit} reports local-return and 8B fallback outcomes as candidate-set size increases. ``Correct'' counts final correct cascade answers, ``Local'' counts local returns, and ``Wrong'' counts errors among those returns. ``Reads'' is the total number of object--attribute reads.

Section~\ref{sec:candidate-breadth} discusses the cascade outcomes in Table~\ref{tab:gqa-tau-audit}. The corresponding local risks are 18.18\%, 11.76\%, and 5.88\% at the three thresholds. Relative to $\tau=0.8$, the accuracy differences are $-2.34$ points with interval $[-6.25,1.56]$ at 0.6 and $-0.78$ points with interval $[-3.91,2.34]$ at 0.95. These paired image-cluster intervals use 10,000 draws on fixed predictions and describe the fixed development conditions. Appendix~\ref{app:protocols} documents the rationale for the primary threshold.

\begin{table}[!t]
\caption{Threshold effects on 128 GQA development questions.}
\label{tab:gqa-tau-audit}
\centering\small
\setlength{\tabcolsep}{5pt}
\begin{tabular}{@{}rrrrrr@{}}\toprule
$\tau$ & Correct & Local & Wrong & 8B calls & Reads\\\midrule
0.60 & 99 & 88 & 16 & 40 & 193\\
0.80 & 102 & 68 & 8 & 60 & 207\\
0.95 & 101 & 34 & 2 & 94 & 235\\
\bottomrule\end{tabular}

\end{table}

\subsection{Read-order analysis}
\label{app:read-order}
The primary GQA holdout offers limited scope for read selection: questions reach 1.55 attribute cells on average, with one cell for 59.83\% of questions and two for 32.68\%. WINE reads 1.26 cells on average, 18.27\% fewer than all reachable cells, and stops before reading them all on 17.17\% of questions, including the 6.38\% answered without any read. Unresolved questions read nearly all available cells, 1.55 of 1.57, before escalation because the cascade imposes no read budget. The synthetic experiments in the main paper evaluate read selection under restricted budgets with larger candidate pools.

On the development panel at $\tau=0.8$, WINE, a fixed order, and Final-set ordering each yield 102 correct answers, 68 local returns, eight wrong local returns, and 60 larger-model calls. Total attribute reads are 207, 223, and 206. WINE saves 16 reads relative to the fixed order, reducing the read count by 7.17\%. It uses one more read than Final-set ordering. The three orders produce identical final correctness outcomes on this fixed development panel.

Synthetic WINE and Final-set in Table~\ref{tab:read-budget} share the executor and singleton stopping rule, scoring all intermediate states or only the final answer set, respectively. For $N$ scene objects, the cap $\lfloor f\,4N\rfloor$ counts the full grid of size, color, material, and shape attributes. Selection uses the current reachable unread pool within this scene-wide budget. Full budget permits early stopping. Read-all acquires candidate sets for all $4N$ scene cells before set-valued execution, so it also changes stopping.

\paragraph{Matched acquisition units.}
The added diagnostic crosses observation unit (cell or crop), budget fraction $f\in\{.25,.5,1\}$, and four policies on the same 1,024-question CLEVR panel and the same fitted 256M attribute heads. One complete bank of fresh crop predictions supplies every arm. The policy observes only selected entries; ground-truth attributes and final answers are absent from policy inputs. All 24,576 trajectories are committed before scoring. The original four WINE/Final-set cell-budget trajectories per question match in final output, singleton status, cell count, and crop count. This is a post-hoc diagnostic on an already reported panel, with settings fixed before these additional runs.

The crop condition reveals all four sets of the selected object atomically and permits $\lfloor fN\rfloor$ such acquisitions. Selection still ranks individual cells using the original hypothetical-singleton score, so this test evaluates the existing controller under bundled observations, rather than a jointly optimized crop policy. Fixed order takes the first eligible cell; random order uses one trajectory with seed $20261110+\mathrm{question\_index}$. Random-policy results condition on this single trajectory per question. No 8B fallback is used. Table~\ref{tab:paired-units-full} reports correct singletons divided by all questions and distinct acquired objects.

\begin{table*}[!t]
\centering\small
\caption{Matched acquisition units: accuracy (\%) and mean crops on 1,024 CLEVR questions.}
\label{tab:paired-units-full}
\begin{tabular}{llrrrrrr}
\toprule
 & & \multicolumn{2}{c}{25\% budget} & \multicolumn{2}{c}{50\% budget} & \multicolumn{2}{c}{100\% budget}\\
\cmidrule(lr){3-4}\cmidrule(lr){5-6}\cmidrule(lr){7-8}
Unit & Policy & Acc. & Crops & Acc. & Crops & Acc. & Crops\\
\midrule
Cell & WINE & 10.55 & 6.14 & 57.62 & 6.58 & 91.41 & 6.58\\
Cell & Final-set & 6.64 & 3.02 & 27.25 & 4.58 & 91.70 & 6.38\\
Cell & Fixed order & 1.27 & 2.26 & 7.91 & 4.00 & 91.21 & 6.43\\
Cell & Random order & 1.76 & 4.79 & 14.94 & 6.41 & 91.41 & 6.67\\
\midrule
Crop & WINE & 0.39 & 1.31 & 2.34 & 3.11 & 90.92 & 6.40\\
Crop & Final-set & 0.59 & 1.31 & 2.93 & 3.11 & 91.21 & 6.33\\
Crop & Fixed order & 0.29 & 1.31 & 2.05 & 3.12 & 91.02 & 6.43\\
Crop & Random order & 0.68 & 1.31 & 1.76 & 3.10 & 91.11 & 6.41\\
\bottomrule
\end{tabular}
\end{table*}

Paired intervals resample the 1,024 distinct images 10,000 times with seed 20261110. At half budgets, changing from cells to crops reduces the WINE-minus-Final-set accuracy contrast by 30.96 points $[27.54,34.38]$. The contrast reflects the joint change in budget and simultaneous attribute revelation on fixed questions and observations. WINE uses more crops than Final-set under half-cell acquisition; paired accuracy-difference intervals include zero at half-crop and full budgets. These exploratory, pointwise intervals condition on one attribute-head fit and the fixed policies. Logged CPU replay durations use six-worker execution and remain separate from the live latency measurements.

\paragraph{Additional backbones and compositions.}
\begin{table}[!t]
\centering\small
\setlength{\tabcolsep}{3pt}
\caption{Accuracy over all questions (\%) under attribute-read caps; no 8B fallback.}
\label{tab:read-budget-backbones}
\begin{tabular*}{\columnwidth}{@{\extracolsep{\fill}}lrrrr@{}}
\toprule
 & \multicolumn{3}{c}{CLEVR ($n=1{,}024$)} & CLOSURE\\
\cmidrule(lr){2-4}\cmidrule(l){5-5}
Read policy & 256M & 500M & 2B & 256M\\
\midrule
\multicolumn{5}{@{}l}{\textit{Half cap: 50\%, at most $2N$ reads}}\\
\textbf{WINE} & 57.62 & 56.15 & 58.98 & 25.89\\
Final-set & 27.25 & 26.37 & 27.93 & 10.04\\
\midrule
\multicolumn{5}{@{}l}{\textit{Full cap: 100\%, at most $4N$ reads}}\\
\textbf{WINE} & 91.41 & 89.84 & 93.65 & 85.16\\
Final-set & 91.70 & 90.43 & 94.53 & 86.16\\
\midrule
\multicolumn{5}{@{}l}{\textit{Exhaustive reference: read all $4N$ cells}}\\
Read-all & 90.62 & 88.67 & 92.87 & 83.71\\
\bottomrule
\end{tabular*}
\end{table}

Table~\ref{tab:read-budget-backbones} gives all cell-budget conditions for the backbone and CLOSURE comparisons in Section~\ref{sec:selection-ablation}. The separate 256-question, three-fit CLEVR panel retains a 24.6--27.0-point advantage per SmolVLM backbone. At the full cap, Final-set is 0.3--1.0 points more accurate, while Read-all is 0.8--1.5 points below WINE. Read-all observes every scene cell before execution, changing both acquisition and stopping. Additional noisy observations can exclude true values and invalidate references, so exhaustive acquisition need not improve accuracy.

A separate historical five-fit diagnostic also favors Final-set at half and three-quarter crop budgets. Its different panel and attribute-head fits provide a qualitative consistency check. The matched-question analysis above identifies the supported acquisition benefit as ordering attribute reads; visual-encoding cost depends on the observation unit. Live cascade cost is evaluated separately.

\subsection{Candidate--annotation compatibility}
Of 270 scored cells, 213 have an in-domain annotation that can be mapped to the candidate vocabulary. In Table~\ref{tab:annotation-compatibility}, ``Cell overlap'' measures overlap with an annotated value among these 213 cells. Only 78 questions have mapped annotations for every scored cell; ``Joint overlap'' requires overlap at all such cells within each question. Both rates increase with the threshold. The mean set size is computed over all 270 scored cells, not only realized reads. Missing annotations, aliases, and potentially multiple attribute values make annotation compatibility the relevant interpretation of these empirical rates. Local correctness remains conditional on the structural, program, and evidence premises of Proposition~\ref{prop:sufficiency}.

\begin{table}[!t]
\caption{Candidate-set size and overlap with mapped GQA annotations.}
\label{tab:annotation-compatibility}
\centering\small
\setlength{\tabcolsep}{3pt}
\begin{tabular*}{\columnwidth}{@{\extracolsep{\fill}}rrrrrr@{}}\toprule
 & & \multicolumn{2}{c}{Cell overlap} & \multicolumn{2}{c}{Joint overlap}\\
\cmidrule(lr){3-4}\cmidrule(lr){5-6}
$\tau$ & \shortstack{Mean\\size} & Count & \% & Count & \%\\\midrule
0.60 & 1.51 & 168/213 & 78.87 & 53/78 & 67.95\\
0.80 & 2.33 & 191/213 & 89.67 & 64/78 & 82.05\\
0.95 & 5.43 & 208/213 & 97.65 & 76/78 & 97.44\\
\bottomrule\end{tabular*}

\end{table}

\subsection{Finite execution checks and failed premises}
Section~\ref{sec:execution-audit} reports the deterministic containment audit. It uses no trained-attribute-head fits and checks a finite set of supported operations; the induction argument is given in Appendix~\ref{app:execution}.

Simple counterexamples identify the error sources covered by the premises of Proposition~\ref{prop:sufficiency}. If the retained color set is $\{\mathrm{red}\}$ but the true color is blue, a singleton can be wrong because the candidate set omitted the truth. A wrong relation can select a different object even with exact attribute sets. An executable program may encode another question: verifying that a blue cup is blue establishes blueness rather than answering whether it is red. A vocabulary that omits the true value likewise excludes the true world. These cases locate failures in candidate retention, structural relations, program semantics, and vocabulary coverage.

The combination of repeated observations also affects truth retention. Intersecting an old set $\{\mathrm{red},\mathrm{blue}\}$ with a new set $\{\mathrm{red}\}$ removes the true value when it is blue. A union preserves old candidates, but still contains only red when both observations retain only red. Evidence retention and evidence correctness therefore require separate evaluation.

\subsection{Scope of the synthetic risk comparison}
The CLEVR threshold sweep characterizes the trade-off between local coverage and local risk. It uses one SmolVLM-256M attribute-head fit and 1,024 questions. As the threshold increases from $\tau=0.6$ to $\tau=0.95$, local coverage falls from 93.55\% to 85.16\%, while local accuracy rises monotonically from 98.54\% to 99.77\%. At $\tau=0.8$, WINE returns 5 wrong answers, compared with 17 for Strict top-1 execution on the same single-fit panel.

The matched-local-coverage comparison evaluates local risk at equal local coverage. This separate five-fit analysis uses post-hoc confidence-ranked rejection after full perception. Matching applies to answer retention after full perception; acquisition costs lie outside the matching criterion.

At a target local coverage of 85\%, each fit retains 870 of 1,024 answers, yielding 84.96\% local coverage. WINE minus Strict top-1 local risk is $-0.713$ percentage points, with interval $[-1.379,-0.161]$. At a target rate of 88\%, each fit retains 901 answers, yielding 87.99\% local coverage; the risk difference is $-0.977$ points with interval $[-1.731,-0.355]$. The bootstrap crosses five attribute-head fits and the 1,024 shared image--question pairs and reselects the ranking in each draw. All 5,000 draws are used at the lower rate. At the higher rate, eight draws have insufficient eligible answers and are omitted, leaving 4,992. The latter interval conditions on eligibility in the retained bootstrap draws. These conditional intervals characterize post-hoc rejection of fixed predictions on the existing panel; acquisition costs and generalization to new test data require separate evaluation.

\subsection{Measured small-model component cost}
A separate pilot measures SmolVLM-256M in FP16 at batch size one on an RTX 3090, with attribute heads on the CPU and only the small VLM resident on the GPU. One fixed set of head weights is reused throughout this pilot, with no new head training. Eight previously exposed images provide 16 timed encodings across original and context views after two warm-up encodings. Mean component time is 58.93\,ms, including 42.84\,ms for the small-VLM forward pass, 12.73\,ms for preprocessing, 1.62\,ms for image decoding, 1.07\,ms for transfer to the GPU, and 0.29\,ms for the CPU attribute head; the remainder is cropping, feature transfer, and postprocessing. This component pilot includes image processing and visual encoding, whereas isolated attribute-head timing measures only the head.

The pilot excludes program prediction, retrieval, the larger model, and the routing loop, and describes component timing on the sampled images. Cascade cost is measured jointly in Appendix~\ref{app:matched-runtime}, which charges perception, routing, and fallback within each question. This common timing boundary avoids combining component times from separate runs; larger-model calls, tokens, resident memory, and latency remain distinct quantities.

\FloatBarrier
\section{Published-component comparisons and input corruption}
\label{app:comparisons}
Supporting component comparisons characterize acceptance and correctness supervision under shared perception. Separate corruption studies perturb the supplied inputs while retaining the original images, questions, and reference answers. The panels reuse historical evaluation data; the fitted attribute heads and stored visual probabilities remain fixed. GQA fallback uses the existing image-and-question-only 8B answers. Table~\ref{tab:published-main} summarizes the two component adaptations; Tables~\ref{tab:published-clevr} and~\ref{tab:published-selector} provide full metrics and intervals. These studies quantify accuracy and routing separately from live timing in Appendix~\ref{app:matched-runtime}.

\subsection{Published components with shared inputs}
\label{app:published}
\subsubsection{Neuro-symbolic execution}
The CLEVR comparison uses the 1,024-question panel, one question per image, and the primary SmolVLM-256M attribute-head fit from Table~\ref{tab:read-budget}. The official NS-VQA executor~\citep{yi2018neural} receives the supplied program and the most likely value of each attribute under the same probability bank. Its relation modules are adapted to consult the supplied relation graph instead of 3D coordinates. The official prefix interpreter and all other operator modules, including the original single-object selection semantics, are retained. This tests the symbolic execution component; the original detector, attribute network, and program parser are outside the comparison.

The Strict top-1 executor executes the same argmax attributes through the concrete executor used by WINE. Both concrete baselines read all $4N$ cells before execution. WINE uses the full cell cap with candidate mass $\tau=0.8$ and may terminate early. A failed or unresolved execution counts as incorrect, with no 8B fallback in this panel. The official relation/prefix adapter returns the annotated answer on all 1,024 questions when supplied gold attributes; this is an implementation diagnostic, separate from learned-perception accuracy.

\begin{table}[!t]
\centering\small
\caption{CLEVR execution comparison with shared perception.}
\label{tab:published-clevr}
\setlength{\tabcolsep}{3pt}
\begin{tabular*}{\columnwidth}{@{\extracolsep{\fill}}lrrrrr@{}}
\toprule
Method & \shortstack{Acc.\\(\%)} & \shortstack{Local\\coverage\\(\%)} & \shortstack{Local risk\\(\%)} & Cells & Crops\\\midrule
WINE & 91.41 & 91.89 & 0.53 & 13.29 & 6.58\\
Strict top-1 & 89.16 & 90.82 & 1.83 & 26.86 & 6.71\\
NS-VQA (adapted) & 91.70 & 93.95 & 2.39 & 26.86 & 6.71\\
\bottomrule\end{tabular*}
\end{table}

Table~\ref{tab:published-clevr} reports accuracy over all questions, local coverage, local risk, and mean observation counts. WINE and the adapted NS-VQA executor reach 91.41\% and 91.70\% accuracy. WINE minus NS-VQA accuracy is $-0.29$ percentage points, with a paired 95\% interval of $[-2.15,1.56]$. WINE returns fewer wrong answers, with local risk 0.53\% versus 2.39\%, and observes 13.29 rather than 26.86 cells. Mean crops remain close at 6.58 and 6.71. Lower local risk accompanies lower local coverage; coverage is unmatched. These supporting diagnostics show how candidate-set sufficiency selects local answers at a lower-coverage operating point. The separate acquisition comparison isolates the benefit of trace-guided ordering with shared stopping. Similar crop counts distinguish the cell savings from visual-encoding cost.

\subsubsection{Selective prediction}
The Reliable VQA Selector~\citep{whitehead2022reliable} comparison uses the original 10,509-question GQA holdout and fixed attribute-head fit 0. All gates share the local-answer array from Section~\ref{sec:setup}: a WINE singleton when available and otherwise the modal answer from 256 sampled executions. They also share the stored 8B answers. Thus the experiment compares routing for this common answerer, rather than independent end-to-end VQA systems. The five-fit Supervised MLP result in Table~\ref{tab:routing-rules} remains a separate comparison.

The original routing-fit images are split by a fixed hash into 4,485 training images and 1,122 validation images, containing 12,480 and 3,234 questions. Holdout contains 3,738 images and 10,509 questions. These image sets are mutually disjoint and have zero overlap with the attribute-head training images. The public input vocabulary includes object names, relation names, choice literals, attribute values, and an invalid-execution entry; it uses no reference answers. Text vocabulary and normalization statistics are fitted only on training questions.

Image features concatenate the mean 576-dimensional frozen crop/name embedding with mean and maximum attribute probabilities, yielding 710 dimensions. Question features concatenate train-fitted TF-IDF features, capped at 512 terms, with counts of 11 operator families, four structure counts, and the identity of the fixed local answer, yielding 2,432 dimensions. Full answer probabilities come from 256 sampled concrete executions with sample seed 7 and sorted cell order. The selected answer may differ from the sampled mode, so its identity is supplied explicitly. The WINE stopping bit and trace features are excluded from the Selector input. The shared answer itself still depends on WINE, as stipulated by the common-answer protocol.

We use the official three-branch Selector network and its two-output correctness BCE objective. Image, question, and answer projections each have 128 dimensions; hidden layers have 256 units with dropout 0.5. The answer branch applies softmax to log probabilities floored at $10^{-6}$. The optimizer is AdamW with learning rate $10^{-4}$, weight decay 0.01, and batch size 256. Each of three training seeds runs 40 epochs, retaining the checkpoint with minimum validation BCE. The inputs and dimensions adapt the published Selector to the shared WINE answerer; the original CLIP-ViL/VQA-v2 training pipeline is excluded.

\begin{table*}[!t]
\centering\small
\caption{Supporting GQA routing comparison at 45.87\% 8B calls with attribute-head fit 0.}
\label{tab:published-selector}
\setlength{\tabcolsep}{3pt}
\begin{tabular*}{\textwidth}{@{\extracolsep{\fill}}lrrrr@{}}
\toprule
Gate & Accuracy (\%) & Local risk (\%) & AURC & Accuracy minus WINE (points) [95\% CI]\\\midrule
WINE & 78.27 & 13.08 & --- & ---\\
Selector seed 1 & 78.26 & 12.88 & 0.1275 & $-0.01\;[-0.53,0.50]$\\
Selector seed 2 & 78.56 & 12.41 & 0.1256 & $0.29\;[-0.22,0.80]$\\
Selector seed 3 & 78.49 & 12.85 & 0.1283 & $0.22\;[-0.27,0.70]$\\
Assigned-answer probability & 76.03 & 18.30 & 0.1787 & $-2.24\;[-2.70,-1.80]$\\
Temperature-scaled probability & 75.46 & 18.47 & 0.1790 & $-2.81\;[-3.29,-2.33]$\\
\bottomrule\end{tabular*}
\end{table*}

The score of the fixed shared answer supplies a probability baseline. A second control temperature-scales the sampled distribution; validation BCE selects from 81 log-spaced temperatures between 0.25 and 4, yielding 1.3660. These scores refer to the assigned shared answer, which need not be the highest-probability answer. They therefore differ from a MaxProb gate that defines its own answer by argmax.

With shared supervised attribute heads, WINE uses trace-based stopping without fitting a separate correctness gate, while the adapted Selector uses additional correctness supervision. Table~\ref{tab:published-selector} retains exactly 5,689 local answers for every gate, leaving 4,820 questions for 8B: a 45.87\% call rate and 54.13\% local coverage. Ranking uses confidence with question-ID tie breaking, without holdout correctness. Selector accuracy ranges from 78.26\% to 78.56\%, with mean 78.43\%, versus 78.27\% for WINE. The reported range is the minimum and maximum over training seeds. Each paired difference interval includes zero; comparable observed values do not establish statistical equivalence. WINE derives acceptance directly from the execution trace, achieving similar observed accuracy at a matched call rate without the additional gate training required by the adapted Selector. The five-fit MLP comparison separately shows the accuracy benefit available from learned routing; each result concerns its specified gate and protocol.

AURC is the mean cumulative error fraction across all retention counts, with lower values preferred. WINE has a fixed operating point rather than a confidence ranking. For validation-selected operating points, we retain the largest complete confidence-tie group whose empirical validation risk meets 1\%, 5\%, or 10\%. The resulting threshold is then fixed on holdout. Table~\ref{tab:selector-thresholds} reports realized local coverage and local risk, including cases with zero local coverage. These empirical targets are not risk guarantees.

\begin{table}[!t]
\centering\small
\caption{Holdout outcomes at validation-selected risk targets.}
\label{tab:selector-thresholds}
\setlength{\tabcolsep}{3pt}
\begin{tabular*}{\columnwidth}{@{\extracolsep{\fill}}lrrrr@{}}
\toprule
Gate & \shortstack{Target\\risk (\%)} & \shortstack{Local cover-\\age (\%)} & \shortstack{Local risk\\(\%)} & \shortstack{Cascade\\acc. (\%)}\\\midrule
Selector 1 & 1 & 4.48 & 2.55 & 77.10\\
Selector 1 & 5 & 16.49 & 4.10 & 78.11\\
Selector 1 & 10 & 42.90 & 9.69 & 78.83\\
Selector 2 & 1 & 4.63 & 2.26 & 77.13\\
Selector 2 & 5 & 17.62 & 4.27 & 78.29\\
Selector 2 & 10 & 43.18 & 9.37 & 78.89\\
Selector 3 & 1 & 3.83 & 2.24 & 76.99\\
Selector 3 & 5 & 17.77 & 4.66 & 78.19\\
Selector 3 & 10 & 42.08 & 9.61 & 78.84\\
Answer prob. & 1 & 0.00 & --- & 76.45\\
Answer prob. & 5 & 0.00 & --- & 76.45\\
Answer prob. & 10 & 10.00 & 8.94 & 76.42\\
Temp. prob. & 1 & 0.00 & --- & 76.45\\
Temp. prob. & 5 & 0.00 & --- & 76.45\\
Temp. prob. & 10 & 10.00 & 8.94 & 76.42\\
\bottomrule\end{tabular*}
\end{table}

\subsection{Controlled scene and program corruption}
\label{app:input-noise}
We vary the fraction $r\in\{0,0.05,0.10,0.20,0.30,0.50\}$ of eligible question inputs receiving a specified edit. For each of three fixed seeds, eligible question IDs are hash-ordered; each rate edits the first $\lfloor rM\rfloor$ entries for $M$ eligible questions. The subsets are nested. Edits are selected without reference answers or predictions of correctness. Images, natural-language questions, reference answers, and visual probabilities remain unchanged. Selected questions receive one relation edit, one program edit, or one of each in the joint condition. A changed program can remain executable and can also leave the correct answer unchanged; membership is based on the edit, not whether the answer changes.

\paragraph{GQA relation errors.}
For each question, eligible edges have a relation name used by a relation, relation-verification, or relation-choice step. One such edge is selected uniformly and deleted. Object names and boxes remain supplied. Editing acts on the question-specific copy of the graph; questions sharing an image need not share the same edited edge. This is a controlled sensitivity test, distinct from the shared predicted graph evaluated in Appendix~\ref{app:structure}.

\paragraph{GQA program errors.}
An eligible step is selected uniformly, followed by a uniform compatible replacement. Attribute filters and verifications exchange literals within their color/material domain, preserving negation. Choice steps replace one alternative with another domain value. Boolean conjunction/disjunction and same/different operators exchange only with their corresponding same-signature alternatives. Dependency indices and attribute categories remain unchanged. The edited reachable-object set is checked to be a subset of the original set, ensuring that no new uncached perception is required. These edits perturb semantic faithfulness while preserving the supported execution grammar.

There are 2,662 eligible relation questions, 6,962 eligible program questions, and 1,455 questions eligible for both. The clean 10,509-question panel is retained in every denominator, and all five existing attribute-head fits are evaluated with each corruption seed. Table~\ref{tab:gqa-corruption} includes both the eligible-set rate and the realized fraction of all questions changed. Accuracy on the full panel and eligible subset are distinguished. Wrong local / all counts incorrect accepted answers over all questions, whereas local risk divides by accepted answers only. Stored 8B predictions are reused for questions routed by each corrupted run; no new VLM timing is inferred from this replay.

\begin{table*}[!t]
\centering\small
\caption{GQA corruption results over five attribute-head fits and three edit seeds.}
\label{tab:gqa-corruption}
\setlength{\tabcolsep}{3pt}
\begin{tabular*}{\textwidth}{@{\extracolsep{\fill}}lrrrrrrr@{}}
\toprule
Input edit & \shortstack{Eligible\\edited (\%)} & \shortstack{Edited\\questions} & \shortstack{All questions\\edited (\%)} & \shortstack{Cascade\\acc. (\%)} & \shortstack{8B routing\\(\%)} & \shortstack{Wrong local /\\all (\%)} & \shortstack{Eligible-set\\acc. (\%)}\\\midrule
Clean & 0 & 0 & 0.00 & 78.29 & 44.50 & 7.38 & ---\\
Relation & 5 & 133 & 1.27 & 78.20 & 44.69 & 7.43 & 78.58\\
Relation & 10 & 266 & 2.53 & 78.09 & 44.87 & 7.48 & 78.14\\
Relation & 20 & 532 & 5.06 & 77.86 & 45.30 & 7.54 & 77.25\\
Relation & 30 & 798 & 7.59 & 77.63 & 45.76 & 7.62 & 76.35\\
Relation & 50 & 1331 & 12.67 & 77.29 & 46.61 & 7.71 & 75.00\\
Program & 5 & 348 & 3.31 & 77.46 & 44.17 & 8.28 & 82.98\\
Program & 10 & 696 & 6.62 & 76.70 & 43.88 & 9.12 & 81.84\\
Program & 20 & 1392 & 13.25 & 75.04 & 43.09 & 10.96 & 79.33\\
Program & 30 & 2088 & 19.87 & 73.56 & 42.43 & 12.61 & 77.09\\
Program & 50 & 3481 & 33.12 & 70.40 & 41.05 & 16.11 & 72.33\\
Joint & 5 & 72 & 0.69 & 78.12 & 44.57 & 7.54 & 89.82\\
Joint & 10 & 145 & 1.38 & 77.95 & 44.68 & 7.68 & 88.64\\
Joint & 20 & 291 & 2.77 & 77.56 & 44.85 & 7.97 & 85.82\\
Joint & 30 & 436 & 4.15 & 77.22 & 45.10 & 8.21 & 83.38\\
Joint & 50 & 727 & 6.92 & 76.62 & 45.43 & 8.71 & 79.00\\
\bottomrule\end{tabular*}
\end{table*}

\begin{table*}[!t]
\centering\small
\caption{CLEVR corruption with fixed perception and no 8B fallback.}
\label{tab:clevr-corruption}
\setlength{\tabcolsep}{3pt}
\begin{tabular*}{\textwidth}{@{\extracolsep{\fill}}lrrrrrrrr@{}}
\toprule
Input edit & \shortstack{Eligible\\edited (\%)} & \shortstack{Edited\\questions} & \shortstack{Realized\\rate (\%)} & \shortstack{WINE\\acc. (\%)} & \shortstack{NS-VQA\\acc. (\%)} & \shortstack{Strict top-1\\acc. (\%)} & \shortstack{WINE\\unresolved (\%)} & \shortstack{WINE wrong\\local / all (\%)}\\\midrule
Clean & 0 & 0 & 0.00 & 91.41 & 91.70 & 89.16 & 8.11 & 0.49\\
Relation & 5 & 51 & 4.98 & 91.18 & 91.28 & 88.87 & 8.20 & 0.62\\
Relation & 10 & 102 & 9.96 & 91.05 & 90.95 & 88.64 & 8.27 & 0.68\\
Relation & 20 & 204 & 19.92 & 90.56 & 90.17 & 87.96 & 8.59 & 0.85\\
Relation & 30 & 307 & 29.98 & 90.17 & 89.39 & 87.40 & 8.76 & 1.07\\
Relation & 50 & 512 & 50.00 & 89.26 & 87.96 & 86.20 & 9.24 & 1.50\\
Program & 5 & 51 & 4.98 & 88.48 & 88.35 & 85.64 & 9.64 & 1.89\\
Program & 10 & 102 & 9.96 & 85.84 & 85.09 & 82.26 & 11.07 & 3.09\\
Program & 20 & 204 & 19.92 & 80.83 & 78.26 & 75.07 & 13.74 & 5.44\\
Program & 30 & 307 & 29.98 & 75.20 & 71.29 & 67.81 & 16.89 & 7.91\\
Program & 50 & 512 & 50.00 & 65.82 & 58.82 & 54.85 & 21.65 & 12.53\\
Joint & 5 & 51 & 4.98 & 88.41 & 88.31 & 85.61 & 9.60 & 1.99\\
Joint & 10 & 102 & 9.96 & 85.81 & 85.03 & 82.26 & 11.00 & 3.19\\
Joint & 20 & 204 & 19.92 & 80.63 & 78.12 & 75.07 & 13.80 & 5.57\\
Joint & 30 & 307 & 29.98 & 74.90 & 71.09 & 67.77 & 16.93 & 8.17\\
Joint & 50 & 512 & 50.00 & 65.30 & 58.46 & 54.72 & 22.07 & 12.63\\
\bottomrule\end{tabular*}
\end{table*}

The clean eligible-subset accuracies are 78.96\% for relations, 84.24\% for programs, and 91.12\% for joint edits. At 20\% eligible program corruption, 1,392 questions change, accounting for 13.25\% of the full panel. Cascade accuracy falls from 78.29\% to 75.04\%, a $-3.25$-point change with interval $[-3.51,-3.01]$. Wrong local / all rises from 7.38\% to 10.96\%, while routing falls from 44.50\% to 43.09\%. The increase in accepted mistakes despite fewer 8B calls identifies semantic corruption that a singleton check cannot detect. At 20\% relation corruption, 532 questions change and accuracy is 77.86\%. Because eligibility, edits, and affected fractions differ, these whole-panel changes do not directly rank the intrinsic severity of relation and program errors.

\paragraph{CLEVR edits.}
All 1,024 questions are eligible in each arm. For relation noise, one existing left/right or front/behind pair is selected and reversed with its inverse entry; no self-edge is introduced. This preserves inverse consistency but need not produce a globally consistent physical scene. Program edits replace a filter literal within its attribute domain, reverse a relation direction, exchange union/intersection, or exchange equal-integer/less-than/greater-than. The program parser verifies operator signatures, the output type, and dependency structure. These are input perturbations, with original images and question labels retained.

Table~\ref{tab:clevr-corruption} reports all three executors with a single fixed attribute-head fit and three edit seeds. CLEVR uses no recorded 8B fallback, so unresolved rate is distinct from measured cascade calls. For WINE, 20\% program editing changes 204 questions and lowers accuracy from 91.41\% to 80.83\%; the accuracy change is $-10.58$ points $[-11.78,-9.41]$. The eligible-set rate is rounded down to an integer count, making the realized rate 19.92\%. Relation-using questions form a 483-question subset; its outcomes and individual seed results are retained in the experiment records. As with GQA, random eligible edits are not selected to alter the answer and should not be interpreted as adversarial or naturally occurring parser-error distributions.

\subsection{Verification and statistical scope}
The official NS-VQA source is pinned to commit \href{https://github.com/kexinyi/ns-vqa/tree/df357618af224723acffb66a17ce3e94298642a7}{\texttt{df357618}}, and Reliable VQA to \href{https://github.com/facebookresearch/reliable_vqa/tree/d9586afe28c6579deb68a93d819d1cc8503c50a5}{\texttt{d9586afe}}. Source and data hashes, prediction files, split definitions, mutation selections, and checkpoints are recorded in the accompanying experiment archive. The Overleaf archive contains document sources; the experiment archive is separate.

Clean GQA runs reproduce the saved answer, acceptance, and read count for every question under all five attribute-head fits. The CLEVR full-cap and half-cap runs reproduce 91.41\% and 57.62\%. Recomputed sampled-answer modes and frequencies match the original records for all 26,223 routing questions. Inference receives no reference answers; holdout labels are excluded from Selector fitting and checkpoint selection. Corruption and Selector predictions are sealed before correctness scoring. Mutation checks verify determinism, input preservation, supported types, inverse-edge consistency where applicable, and reachable-cell containment. Input hashes are rechecked before analysis.

Additional-study intervals use 2,000 paired bootstrap draws over image clusters, conditional on the fixed fitted heads and edit seeds. They are pointwise percentile intervals without multiplicity adjustment and differ from the crossed-fit 10,000-draw protocol used for the primary routing results. Mutation-seed variation is retained separately from attribute-head-fit variation. The supplied panels have been evaluated previously, so these results extend analysis on historical data rather than establish performance on a new untouched test set. Full detector/parser reproduction and natural input-error distributions remain separate evaluation settings.

\end{document}